\documentclass[10pt,twocolumn,letterpaper]{article}

\usepackage{iccv}
\usepackage{times}
\usepackage{epsfig}
\usepackage{graphicx}
\usepackage{graphics}

\usepackage{amsmath}
\usepackage{amssymb}

\usepackage[pagebackref=true,breaklinks=true,letterpaper=true,colorlinks,bookmarks=false]{hyperref}

\iccvfinalcopy 

\def\iccvPaperID{4072} 

\ificcvfinal\fi

\begin{document}

\title{A Confidence-based Iterative Solver of Depths and Surface Normals \\ for Deep Multi-view Stereo}

\author{Wang Zhao$^{1,3}\thanks{Equal contribution}$ \quad Shaohui Liu$^{2,3*}$ \quad Yi Wei$^{1}$ \quad Hengkai Guo$^{3}$ \quad Yong-Jin Liu$^{1,4}$\\
$^1$Tsinghua University \quad $^2$ ETH Zurich \quad $^3$ ByteDance Inc. \quad $^4$ JCMV \\
\tt\small b1ueber2y@gmail.com, \{zhao-w19, y-wei19\}@mails.tsinghua.edu.cn, \\
\tt\small guohengkai@bytedance.com, liuyongjin@tsinghua.edu.cn
}

\maketitle
\ificcvfinal\thispagestyle{empty}\fi

\begin{abstract}
   In this paper, we introduce a deep multi-view stereo (MVS) system that jointly predicts depths, surface normals and per-view confidence maps. The key to our approach is a novel solver that iteratively solves for per-view depth map and normal map by optimizing an energy potential based on the locally planar assumption. Specifically, the algorithm updates depth map by propagating from neighboring pixels with slanted planes, and updates normal map with local probabilistic plane fitting. Both two steps are monitored by a customized confidence map. This solver is not only effective as a post-processing tool for plane-based depth refinement and completion, but also differentiable such that it can be efficiently integrated into deep learning pipelines. Our multi-view stereo system employs multiple optimization steps of the solver over the initial prediction of depths and surface normals. The whole system can be trained end-to-end, decoupling the challenging problem of matching pixels within poorly textured regions from the cost-volume based neural network. Experimental results on ScanNet and RGB-D Scenes V2 demonstrate state-of-the-art performance of the proposed deep MVS system on multi-view depth estimation, with our proposed solver consistently improving the depth quality over both conventional and deep learning based MVS pipelines. Code is available at \href{https://github.com/thuzhaowang/idn-solver}{\color{cyan}{https://github.com/thuzhaowang/idn-solver}}.
\end{abstract}

\vspace{-10pt}
\section{Introduction}

Dense multi-view stereo (MVS) has been a long-standing fundamental topic in computer vision. The key idea of most existing techniques is to compare the similarity of image patches at different depth hypotheses, densely matching pixels across images. While great improvement has been witnessed over decades, it is still a hard problem to accurately estimate dense geometry from posed images in many real-world scenarios, especially in indoor environments, where one of the most typical reasons of failure is the existence of texture-less areas (e.g. walls), which incur significant ambiguities to the matching step, as a number of different depths all result in low matching costs. 

\begin{figure}[tb]
\includegraphics[width=1.0\linewidth]{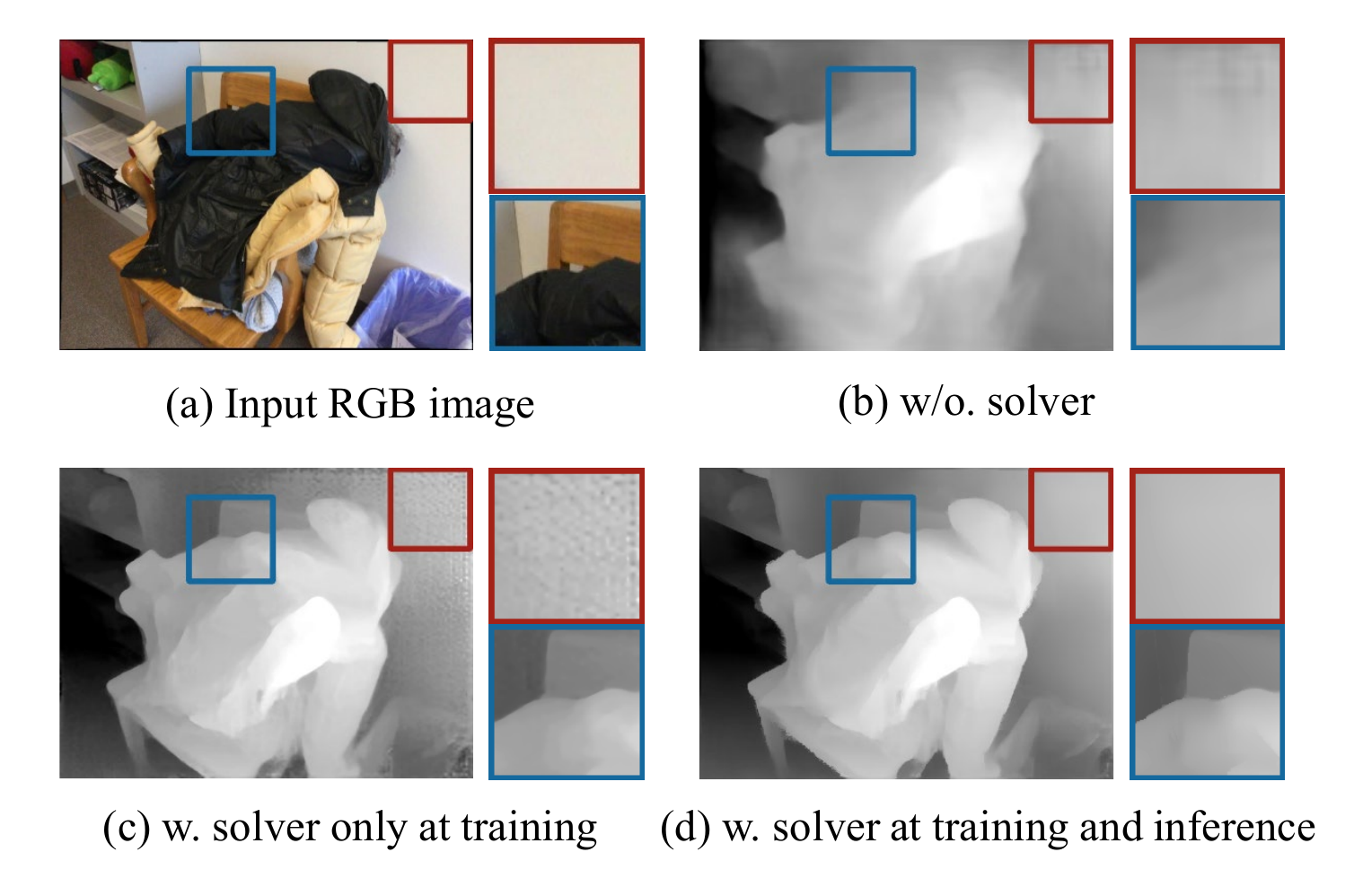}
\centering
\caption{By integrating the proposed iterative solver into end-to-end training, our method decouples the challenging problem of estimating depth values within poorly textured regions from the initial network predictions, getting the network to focus more on reliable estimation on well textured pixels. This improves depth quality on both textured (blue) and texture-less (red) regions. Note that on texture-less areas (red box on (c)(d)), the network jointly trained with the solver only focuses on predicting reliable geometry on neighboring areas, leaving the poorly textured pixels to be resolved with the iterative solver module.}
\label{fig::teaser}
\vspace{-10pt}
\end{figure}

There have been various attempts to tackle this crucial problem. Global optimization \cite{hong2004segment,bleyer2005layered,yamaguchi2012continuous} is one possible direction to resolve the ambiguities. Those methods initially assume a fixed set of superpixels and apply Markov random field with slanted-plane models. The optimization prefers globally smoothed geometry with locally planar surfaces and achieves promising performance. However, global optimization methods suffer from large computational complexity and the performance is also limited by the quality of superpixels and handcrafted dense features. 

With the recent success of deep learning, learning-based methods have achieved great performance on datasets due to the improved quality of learned features and depth priors. Typically, a cost-volume based architecture \cite{yao2018mvsnet,wang2018mvdepthnet} is trained end-to-end with a pixelwise loss function on depths. While the ambiguities incurred by the texture-less areas still exist in the built cost volumes, several methods propose to utilize surface normals, either jointly predicted \cite{qi2018geonet,kusupati2020normal} or online fitted \cite{yin2019enforcing,long2020occlusion}, to help train the depth predictions, aiming to implicitly enrich the network with locally planar priors by directly supervising the local structure of the predicted depth map. However, it is still a difficult problem to successfully learn such priors and accurately predict depths on texture-less areas directly from the ambiguous cost volumes built on warped patches. 

In this paper, we propose a novel deep multi-view stereo system that decouples the locally planar priors from the cost volume based depth/normal prediction networks. The key to our proposed system is a differentiable confidence-based solver that iteratively solves for depth map and normal map by optimizing an energy potential that prefers locally planar surfaces. For each optimization iteration, the solver updates depth map by propagating from neighboring pixels with slanted planes, and updates normal map with local probabilistic plane fitting. A customized confidence map can be used to monitor both two steps. 

Our deep MVS system firstly predicts depth, normal and confidence jointly and then applies multiple optimization steps of our depth-normal solver upon the initial predictions. Those multiple steps enable long-range propagation of reliable depths and normals. The whole system can be trained end-to-end, guiding the depth/normal network to focus more on confident depth/normal predictions on textured areas and leave the poorly textured regions to be resolved with the solver. For the solver module, a hybrid confidence map combining the deep confidence prediction and the conventional geometric reprojection check can be used to stabilize the optimization process at inference.

The proposed system combines the advantages of slanted-plane models \cite{birchfield1999multiway} and learning-based techniques. With end-to-end training, the initial depth prediction is guided to focus only on reliable estimation over partial regions rather than the whole image (as shown in Fig. \ref{fig::teaser}). Experimental results on ScanNet and RGB-D Scenes V2 show that our novel deep MVS achieves state-of-the-art performance in terms of accuracy on depths and surface normals, with the proposed solver consistently improving the depth quality over both conventional and deep MVS pipelines.

\section{Related Work}

\paragraph{Multi-view Stereo.}
Reconstructing 3D models from posed images has been widely studied over decades. Early methods \cite{kutulakos2000theory, faugeras2002variational,vogiatzis2005multi} utilize volumetric optimization. Later attempts \cite{yoon2006adaptive,gallup2007real,bleyer2011patchmatch,hosni2012fast,schonberger2016pixelwise} reconstruct per-view depth maps by comparing cross-view image patches, which becomes the de facto approach of modern multi-view stereo pipelines. Recent advances include employing superpixels \cite{romanoni2019tapa}, post-refinement \cite{kuhn2019plane,kuhn2020deepc}, and advanced propagation checkerboards \cite{galliani2015massively,xu2019multi,xu2020planar}, etc. With the success of deep learning, a number of learning-based techniques are proposed to tackle the problem. While several methods learn to directly predict 3D geometry as grids \cite{kar2017learning,ji2017surfacenet}, point clouds \cite{chen2019point} and TSDF \cite{murez2020atlas}, per-view depth map estimation is still the top choice of most approaches \cite{wang2018mvdepthnet,yao2018mvsnet,im2019dpsnet,liu2019neural,hou2019multi,kusupati2020normal,long2020occlusion,duzcceker2020deepvideomvs,patil2020don} due to its robustness and flexibility. Most of those methods follow the spirit of conventional approaches \cite{gallup2007real,bleyer2011patchmatch} and train a cost volume based neural network. Some of the methods \cite{hou2019multi,duzcceker2020deepvideomvs,patil2020don} also integrate temporal information with recurrent networks. However, direct prediction over poorly textured areas have been one of the main difficulties of most approaches relying on cost volumes. An exception of this is DELTAS \cite{sinha2020deltas}, which proposes to learn interest points and perform triangulation and densification over sparse points. In this work, we also focus on per-view depth estimation and introduce an end-to-end MVS system equipped with a novel iterative solver, which implicitly decouples depth prediction over texture-less areas from the cost-volume based network. 

\vspace{-10pt}
\paragraph{Confidence Estimation.}
Accurately estimating per-view depth map confidence is beneficial to 3D reconstruction pipelines.
Early confidence measures such as matching cost, peak ratio and maximum likelihood are extensively studied in \cite{hirschmuller2008evaluation,hu2012quantitative}. Local smoothness can also contribute to the confidence measurement of the stereo problem \cite{sun2003stereo, hirschmuller2007stereo}. Recently, learning-based methods \cite{seki2016patch,poggi2016learning} are proved to be effective on confidence estimation. Most related to us, \cite{kuhn2020deepc} exploits the counter map acquired from the geometric consistency check for deep confidence prediction. In our MVS system, we combine deep confidence estimation and geometric consistency check together to get our final confidence map, which monitors the solver at inference. 

\vspace{-10pt}
\paragraph{Depth-Normal Constraints.}
As depths and surface normals are naturally coupled by local plane fitting, joint depth-normal constraints are widely employed. Markov random fields with slanted-plane models were initially introduced in \cite{birchfield1999multiway}, and were later used in many stereo techniques \cite{hong2004segment,bleyer2005layered,yamaguchi2012continuous}. Those methods operate on superpixels and prefer locally planar surfaces. \cite{bleyer2011patchmatch} proposed to employ slanted support windows in its PatchMatch stereo framework, coupling the depth propagation with the randomly initialized per-pixel surface normal. \cite{graber2015efficient} further introduce a minimal-surface regularization method. \cite{valentin2018depth} propose to embed bilateral filtering in the plane fitting process to encourage locally planar structures of the output depth map. Recent learning-based methods also exploit surface normals to help train the depth prediction. While \cite{eigen2015predicting,qi2018geonet,kusupati2020normal} jointly predict depths and surface normals together to benefit from multi-task feature learning, recent literature \cite{yin2019enforcing,long2020occlusion} adopts online fitting. In \cite{zhang2018deep}, a surface normal network is trained to help solve for dense depth map from sparse depth observations. Different from prior works, our method explicitly exploits depth-normal constraints via a differentiable iterative solver.

\section{An Iterative Depth-Normal Solver}
The key to our deep multi-view stereo system is a confidence-based iterative depth-normal solver module, which solves for per-view depth map and normal map from the initial predictions. Specifically, we assume locally planar structure of the depth map and couple the normal map within our optimization scheme. 

\subsection{Energy Formulation}
Our energy potential $E_{total}$ consists of a data term $E_{data}$ and a plane-based structural term $E_{plane}$, with the hyperparameter $\alpha$ weighting the two terms: 

\begin{equation}
    E_{total} = \alpha E_{data} + E_{plane},
    \label{eq::energy}
\end{equation}

The data term regularizes the output geometry with respect to the given initial geometry. Take $d_i$, $n_i$ as the per-pixel depth and surface normal respectively, and $c_i$ as the per-pixel confidence. Denote the initial depth and surface normal as $\hat{d}_i$ and $\hat{n}_i$. The data term is written as follows:

\vspace{-5pt}
\begin{equation}
    E_{data} = \sum_i c_i(d_i - \hat{d}_i)^2 + \sum_i c_i||n_i - \hat{n}_i||^2.
    \label{eq::dataterm}
\end{equation}
\vspace{-5pt}

The plane-based structural term $E_{plane}$ enforces the output depth and surface normal to agree with the local planar assumption with respect to neighboring geometry. Let $(x_i, d_i, n_i)$ and $(x_j, d_j, n_j)$ denote the 2D coordinate, depth and surface normal of the current pixel and its neighboring pixel respectively. $P(x, d, n)$ denotes the slanted plane at position $x$ generated by spanning a plane from the corresponding 3D points recovered from $d$ and $x$ with the surface normal $n$. Our plane-based structural energy can be formulated in either of the two below directions: 
\begin{itemize}
    \item $E_{j \rightarrow i}$. The sum (over each pixel) of distance functions between the 3D point recovered from $d_i$ and $x_i$ at the current pixel $i$ and the slanted plane $P(x_j, d_j, n_j)$ generated at its neighboring pixel $j$. $d_{j\rightarrow i}$ denotes the projection of the slanted plane $P(x_j, d_j, n_j)$ at pixel $i$.
    \item $E_{i \rightarrow j}$. The sum (over each pixel) of distance functions between the slanted plane $P(x_i, d_i, n_i)$ generated at the current pixel $i$ and the 3D point recovered from $x_j$ and $d_j$ at its neighboring pixel $j$. $d_{i\rightarrow j}$ denotes the projection of the slanted plane $P(x_i, d_i, n_i)$ at pixel $j$. 
\end{itemize}

\begin{figure}[tb]
\scriptsize
\setlength\tabcolsep{1.0pt} 
\begin{tabular}{ccc}
{\includegraphics[width=0.33\linewidth]{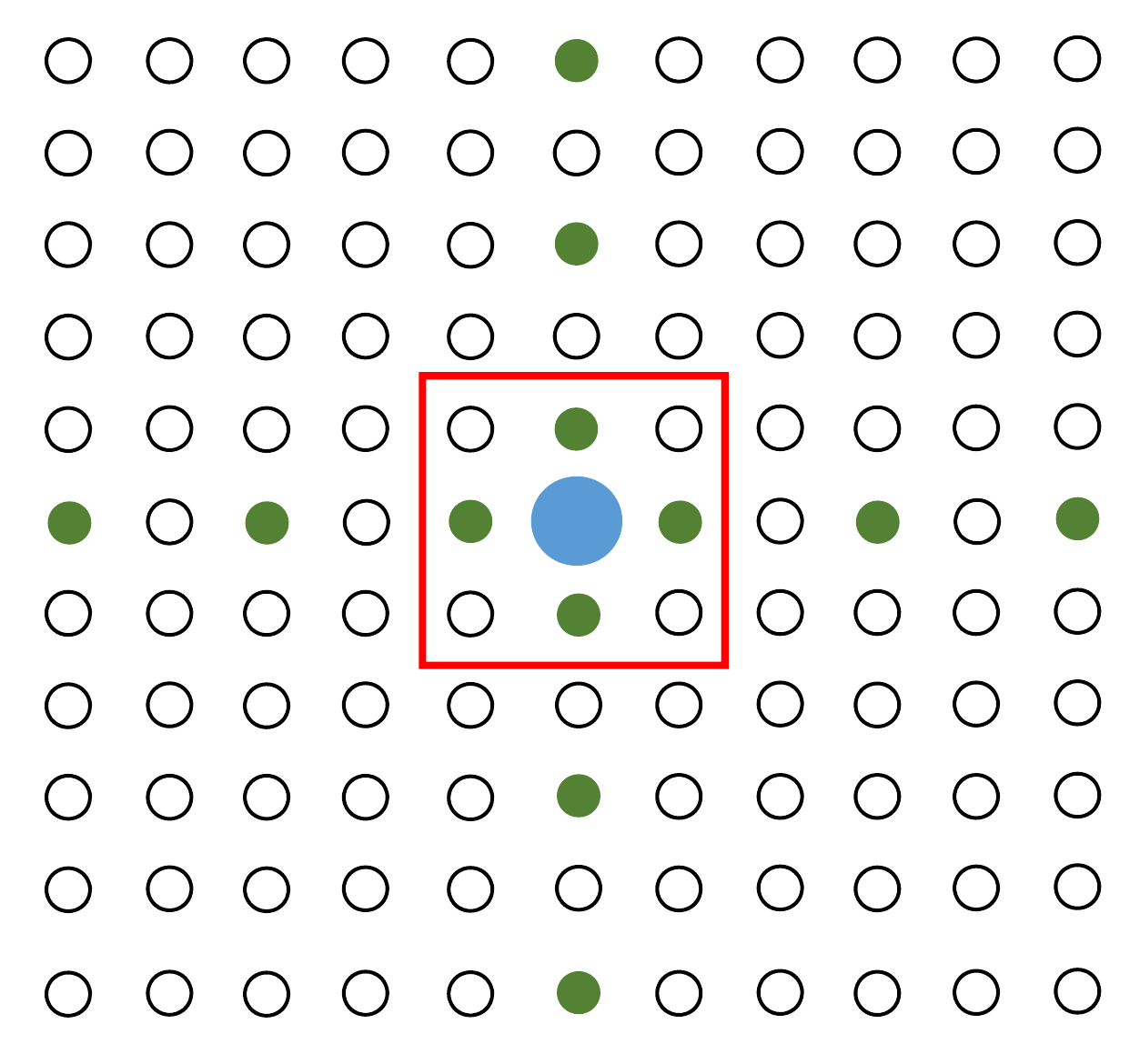}} & 
{\includegraphics[width=0.33\linewidth]{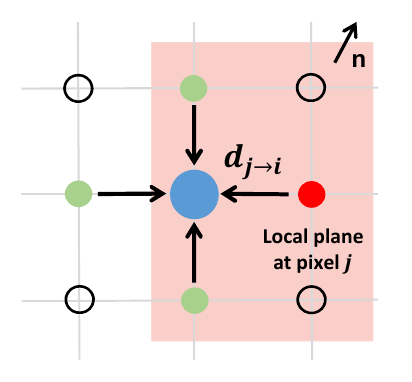}} &
{\includegraphics[width=0.33\linewidth]{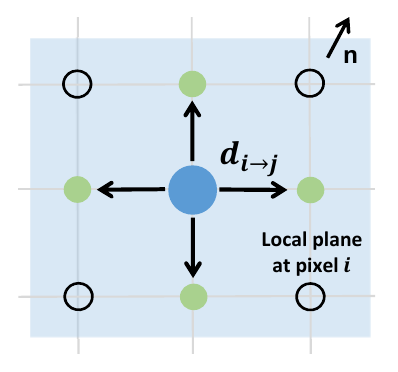}}
 \\
 (a) Checkerboard & (b) D-step & (c) N-step \\

\end{tabular}
\centering
\vspace{0.05cm}
\caption{Illustration of the proposed solver. (a) we use a sparse checkerboard to define the neighborhood of each pixel, which enables large receptive fields and efficient computation. (b) In the D-step, we update depth map by propagating from neighboring pixels with slanted planes. (c) In the N-step, we update normal map with local probabilistic plane fitting. }
\label{fig::solver}
\vspace{-10pt}
\end{figure}

For the definition of neighboring pixels, 
we can either use a sparse checkerboard \cite{galliani2015massively,xu2019multi} (as in Fig. \ref{fig::solver}(a)) or randomly sample sparse points within a fixed window. This enables efficient long-range propagation of the neighboring geometry. In both formulations of the structural energy the contribution of each pixel is also monitored by the per-pixel confidence $c_i$. Motivated by the effective formulation in the bilateral filtering techniques \cite{barron2015fast,barron2016fast}, we also weight the contribution of the neighboring pixels with the edge-aware bilateral affinity, denoted as $w_{ij}$:

\begin{equation}
    w_{ij} = \exp(-\frac{||x_i - x_j||^2}{2\sigma_x^2}-\frac{||I_i - I_j||^2}{2\sigma_c^2}),
\end{equation}
where $I_i$ denotes the RGB value at each pixel. While we use RGB colorspace here for simplicity, the weight can be easily extended to YUV space, as employed in \cite{barron2015fast,barron2016fast}.  

\begin{figure*}[tb]
\includegraphics[height=230pt, width=\linewidth]{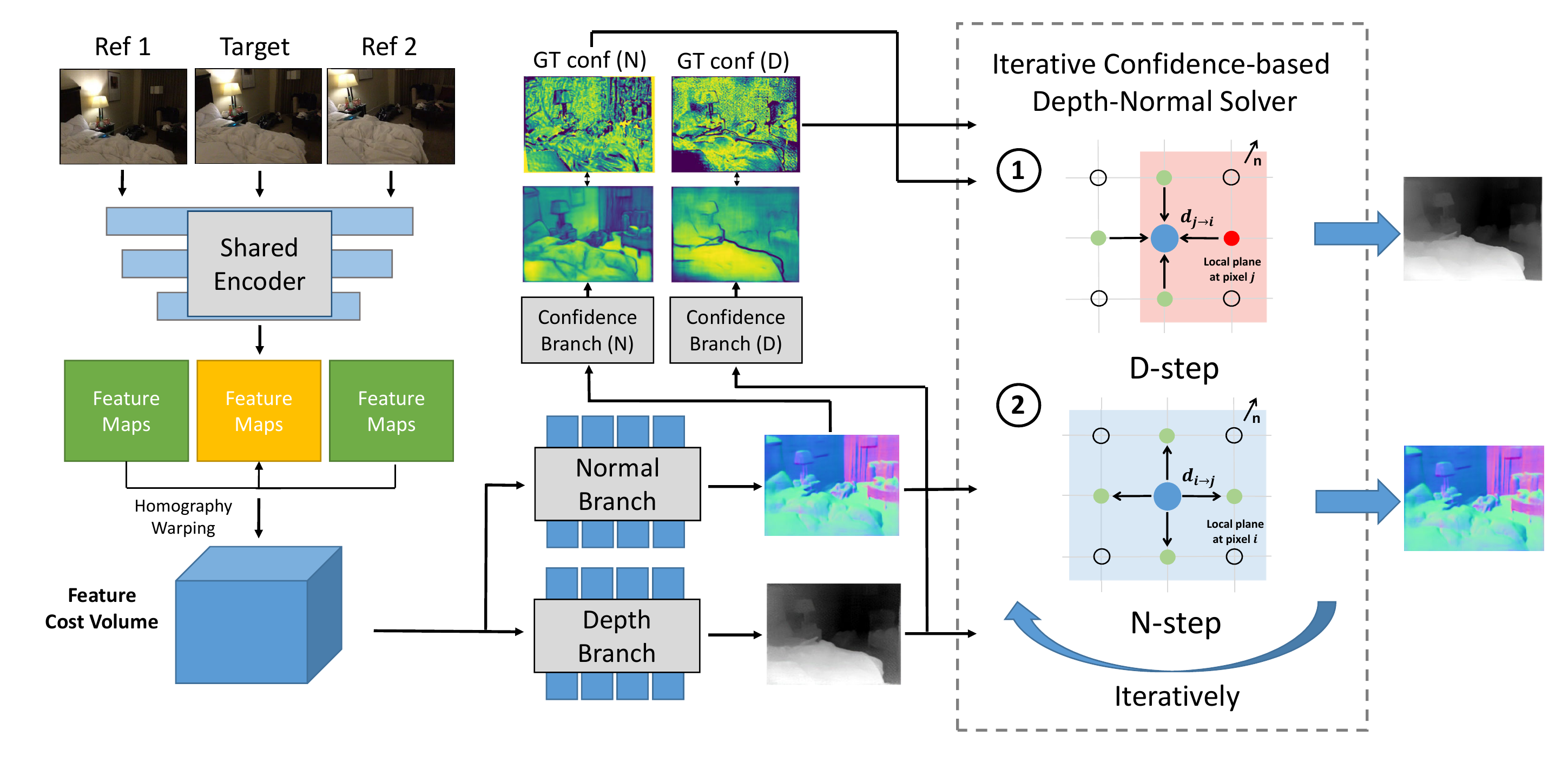}
\centering
\caption{An overview of our deep multi-view stereo (MVS) system. First, We feed the target image and multiple reference images into a shared encoder to extract per-view deep image features, which are used to build a feature cost volume. Then, we jointly predict depth, surface normal and confidence from the cost volume with 3D CNNs and 2D dilated CNNs \cite{yu2017dilated}. Finally, the confidence-based depth-normal solver is applied over the predicted depths and surface normals iteratively to get the final output. The whole system can be trained end-to-end. At training, a groundtruth confidence map computed by the relative depth error is used to help guide the initial prediction to focus on estimating reliable geometry on textured areas. Note that the inputs of the confidence branch are simplified for better visualization. }
\label{fig::pipeline}
\vspace{-10pt}
\end{figure*}

\subsection{Iterative Subproblem Optimization of Depths and Surface Normals}

We aim to minimize the total energy with an effective, parallelized and differentiable approximation. To parallelize the computation, we assume fixed neighboring geometry and solve for each pixel individually at each update step. Since depths and surface normals are non-linearly coupled when propagating from neighboring pixels due to the perspective projection of the slanted planes, closed-form solution can not be acquired for the whole energy potential with depths and normals both considered in the data term\footnote{If only the surface normal data term is considered, closed-form solution can be derived by substitution of variables.}. 

Our proposal is to iteratively solve for depths and surface normals. As shown in Figure \ref{fig::solver}(b)(c), Each iteration is further divided into two steps of subproblem optimization, where we solve for depths/normals individually with the other variables (normals/depths) fixed. Those two sub-problems can all have closed-form solutions. Furthermore, we alternate between the two plane-based structural term formulations to better fit each sub-problem, acquiring a decoupled linear system algebraically for both steps, thus making the gradients of the whole solver tractable. 

\paragraph{Depth Update (D-step).} In the depth update step, we fix the surface normal map and solve for the optimal depth map that minimizes the depth energy $E_d$ as follows:
\vspace{-2pt}
\begin{align}
    & \min_d E_{total} = \min_d E_{d} \\
    & E_{d} = \alpha \sum_i c_i(d_i - \hat{d}_i)^2 + \sum_i \sum_{j \in N(i)}c_jw_{ij}(d_i - d_{j\rightarrow i})^2,
\end{align}

where $N(i)$ denotes the defined neighborhoods of the pixel $i$. We employ $E_{j \rightarrow i}$ as the plane-based structural term and computes the L2 distance between the optimized depth and the propagated depth $d_{j \rightarrow i}$. As previously discussed, we assume fixed neighboring geometry, so the propagated depth $d_{j \rightarrow i}$ here is the projection of the slanted plane $P(x_j, \hat{d}_j, \hat{n}_j)$ at pixel i. Compared to $E_{i \rightarrow j}$  which only uses one surface normal query at the pixel itself, employing $E_{j \rightarrow i}$ improves robustness to outliers in the initial surface normal map, enabling the solver to utilize surface normals in all neighboring pixels. We can derive closed-form optimal depth by setting the first-order derivative of $E_d$ to zero.


\paragraph{Surface Normal Update (N-step).} In the surface normal update step, we fix the depth map and solve for the optimal surface normal that minimizes the surface normal energy $E_n$ as follows:
\vspace{-5pt}
\begin{align}
    & \min_n E_{total} = \min_n E_{n} \\
    &
    \begin{aligned}
    E_{n} &= \alpha \sum_i c_i||n_i - \hat{n}_i||^2 \\
    &+ \sum_i \sum_{j \in N(i)} c_jw_{ij}D_n(d_j, P(x_i, d_i, n_i)).
    \end{aligned}
\end{align}
\vspace{-10pt}

$D_n$ is a distance function between the depth $d_j$ at neighboring pixels and the slanted plane $P(x_i, d_i, n_i)$ at pixel i. The plane equation is computed from the depth $d_i$ updated in the last step and the surface normal $n_i$ being optimized, forming a local probabilistic plane fitting problem. Note that simply employing L2 distance between $d_j$ and $d_{i\rightarrow j}$ as in the D-step will result in a non-quadratic surface normal energy due to the perspective projection during plane-based propagation. Instead, we can parameterize the surface normal as $n=(a, b, -1)$ and utilize the algebraic form of the plane equation to acquire a quadratic energy, where 2 linear equations can be acquired by setting the first-order derivatives to zero, being sufficient to solve the 2-DoF surface normal $n_i$ in closed form\footnote{Please refer to our supplementary material for more details.}.

\section{Deep Multi-view Stereo System}
Based on the confidence-based iterative solver, we propose a deep multi-view stereo system that jointly predicts per-view depths, surface normals and confidence maps. Figure \ref{fig::pipeline} shows an overview of the proposed system.

\subsection{Estimating Initial Depths and Surface Normals}
As our work does not focus on the network architectures, we mostly follow prior works \cite{im2019dpsnet,kusupati2020normal} to build the cost-volume based multi-view depth-normal network. First, the target image and multiple reference images are fed into a shared neural network encoder respectively to acquire per-view deep image features. Then, we apply plane sweeping on feature maps to build a feature cost volume. 3D CNNs and 2D dilated CNNs \cite{yu2017dilated} are applied over the cost volume to aggregate and regularize the cost information. The depth maps are then regressed using the soft argmin operator from the final cost volume. The normal branch follows the design of \cite{kusupati2020normal}, which concatenates the world coordinate volume with the feature cost volume, and uses cost volume slices to estimate the surface normal. 

For the confidence estimation, multiple sources are utilized to facilitate the effective confidence regression, including intermediate feature maps, cost volumes, and the predicted depths/normals. We feed them into two mini-networks which consist of several dilated convolutions and a final sigmoid activation, to estimate the confidence for depth and normal respectively. 
Please refer to our supplementary material for details on the network architecture. 

To train the network, we use smoothed L1 loss for depth and normal, and cross-entropy loss for confidence. The groundtruth confidence maps $c_{dgt}$ and $c_{ngt}$ are computed from the relative depth error $e_{rel}$ and normal angle error $e_{ang}$: $c_{dgt} = \max(1.0 - \gamma_1 e_{rel}, 0.0)$ and $c_{ngt} = \max(1.0 - \gamma_2 e_{ang}, 0.0)$, where $\gamma_1$, $\gamma_2$ are hyperparameters. 

\subsection{Integrating the Proposed Solver}
After acquiring the initial predictions, we employ our solver module over the initial depth map and surface normal map. Specifically, D-step and N-step are applied iteratively for multiple iterations. We use 5 iterations for training and more steps can be used at inference. Multiple iterations lead to more accurate approximation of the optimal depth and surface normal under the local planar assumption. Moreover, since the plane-based propagation is limited by the window size of the checkerboard, using multiple iterations enables long-range propagation from reliable pixels. Empirically we find that improving the number of iterations at inference does lead to better depth quality (See Table \ref{tab::ablation-iter-view}).

Since all operations in the solver are differentiable, the whole system can be trained end-to-end, where the loss on the final solved geometry can be back-propagated into the network for training initial depth/normal predictions. At the training stage, we apply the depth loss and normal loss both on the initial predictions and the final solved geometry. 

The end-to-end training of the system brings up several advantages. 
First, the solver can be considered as a closed-form refinement step to the initial depth/normal predictions, which improves robustness to noise and outliers. 
Moreover, the plane-based structural term favors locally planar surfaces, which is particularly beneficial on poorly textured regions and occluded areas in indoor environments, where cost-volume based approaches struggle to estimate accurate geometry due to large ambiguity. 
Integrating the proposed solver into our end-to-end deep MVS system eliminates the necessity for the network to handle pixels in extremely texture-less regions, since it can be left to the propagation steps of the solver at inference. This promotes our network to only focus on reliable estimation on textured areas, which largely releases the learning burden. 
As shown in Figure \ref{fig::teaser}, this mechanism improves the depth quality significantly on both textured and texture-less regions.

To better enrich the network with such behaviors, at training we apply the iterative solver with the groundtruth depth/normal confidence map, which is computed by comparing the error of the depth/normal map with the groundtruth depth/normal. We empirically observe that high-quality confidence map for the solver is needed during training.
Only using the jointly predicted confidence map at training cannot successfully enrich the initial predictions with the aforementioned nice property. 

\subsection{Inference}
\begin{figure}[tb]
\scriptsize
\setlength\tabcolsep{1.0pt} 
\begin{tabular}{cccc}
{\includegraphics[width=0.24\linewidth]{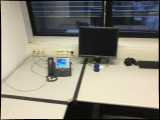}} & 
{\includegraphics[width=0.24\linewidth]{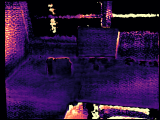}} &
{\includegraphics[width=0.24\linewidth]{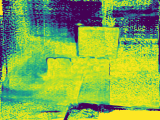}} &
{\includegraphics[width=0.24\linewidth]{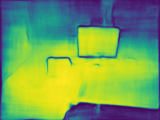}}
 \\
 {\includegraphics[width=0.24\linewidth]{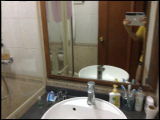}} & 
{\includegraphics[width=0.24\linewidth]{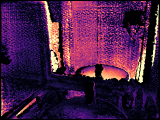}} &
{\includegraphics[width=0.24\linewidth]{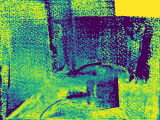}} &
{\includegraphics[width=0.24\linewidth]{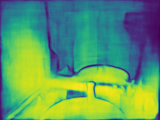}}
 \\
(a) Image & (b) Depth Error & (c) Geometric Conf & (d) Deep Conf\\

\end{tabular}
\vspace{0.05cm}
\centering
\caption{Visualization on different confidence maps. The relative depth error (b) is used to compute the groundtruth confidence. While the confidence acquired by geometric consistency check (c) is more accurate, the deep confidence (d) is complete and acts as a supplements to the geometric confidence at inference.}
\label{fig::confidence}
\vspace{-10pt}
\end{figure}
At inference, we use the trained network to predict initial depths, surface normals and confidence maps. The predicted confidence can be used in the iterative solver with the initial depth map and normal map as input. To further improve the quality of the depth confidence, we also utilize a cross-view geometric consistency check over the predicted depth maps at multiple views to acquire a geometric confidence. The target depth is reprojected and compared with the reference depths to compute the relative depth error, which is then converted into geometric confidence. If multiple reference images are available, we take the minimum over computed confidence maps to get a ``strict" confidence, which reduces the number of false-positive high confidence pixels. As shown in Figure \ref{fig::confidence}, 
while the geometric confidence is often more accurate, the deep confidence is complete under occlusions and small overlaps, which can serve as a good supplements in practice. 
A hybrid confidence map can be acquired by multiplication of the deep confidence and the geometric confidence, which is used in the iterative solver to produce the final output geometry.

\section{Experiments}
\subsection{Implementation Details}
We use the ScanNet dataset \cite{dai2017scannet} to train our system. The official training split is adopted. Three views from one sequence with a fixed frame interval 20 forms a training data sample. The initial depth and normal network is firstly trained with 15 epochs, then integrated with the solver and trained end-to-end for another 10 epochs. We use the Adam optimizer \cite{kingma2014adam} with learning rate 1e-4 and batch size 24 on 4 Nvidia V100 GPUs. For the solver, we define the neighborhood by offsetting each coordinate with 1, 3, 5, 10 pixels horizontally and vertically, forming 16 valid entries of the propagation checkerboard (as illustrated in Figure \ref{fig::solver}(a)). 

\noindent
\textbf{Time efficiency. } A direct implementation in PyTorch takes 14ms for each iteration on a Nvidia V100 GPU. Implementing a CUDA C++ kernel can lead to 0.9ms/iter (15x speed-up) over a 4-Megapixel image, which is nearly negligible compared to the the backbone network.


\subsection{Validating the Solver Module}
\begin{figure}[tb]
\scriptsize
\setlength\tabcolsep{1.0pt} 
\begin{tabular}{ccccc}
{\includegraphics[width=0.19\linewidth]{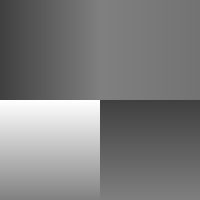}} & 
{\includegraphics[width=0.19\linewidth]{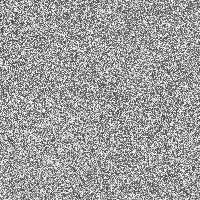}} &
{\includegraphics[width=0.19\linewidth]{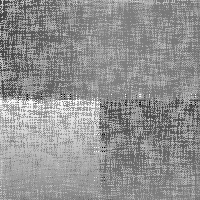}} &
{\includegraphics[width=0.19\linewidth]{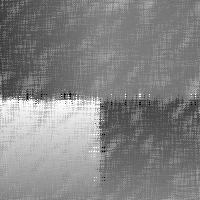}} &
{\includegraphics[width=0.19\linewidth]{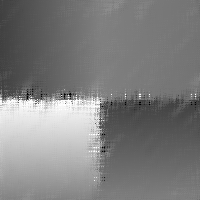}}
 \\

(a) Target & (b) Initial & (c) Iter 1 & (d) Iter 3 & (e) Iter 10 \\

\end{tabular}
\centering
\vspace{0.05cm}
\caption{Visualization on the optimization process of the proposed depth-normal solver for a synthetic proof-of-concept experiment. }
\label{fig::toy}
\vspace{-10pt}
\end{figure}
\begin{figure}[tb]
\scriptsize
\setlength\tabcolsep{1.0pt} 
\begin{tabular}{cc}
{\includegraphics[width=0.46\linewidth]{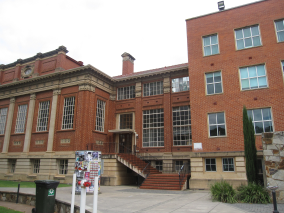}} & 
{\includegraphics[width=0.46\linewidth]{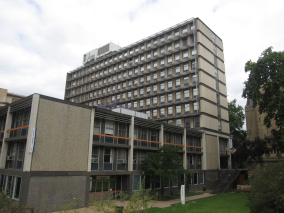}} \\
{\includegraphics[width=0.46\linewidth]{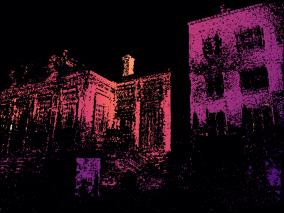}} & 
{\includegraphics[width=0.46\linewidth]{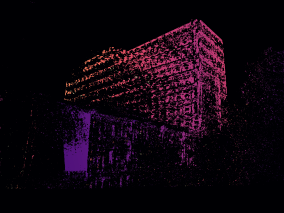}} \\
{\includegraphics[width=0.46\linewidth]{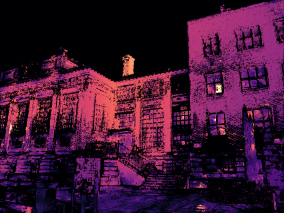}} & 
{\includegraphics[width=0.46\linewidth]{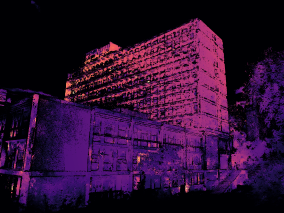}} \\
\end{tabular}
\vspace{0.05cm}
\centering
\caption{Results on applying the proposed solver over sparse reconstructions from COLMAP \cite{schonberger2016structure,schonberger2016pixelwise}. \textbf{Top row:} input image. \textbf{Middle row:} initial sparse depth map acquired from COLMAP. \textbf{Bottom row:} output depth after applying the proposed solver.}
\label{fig::colmap}
\vspace{-10pt}
\end{figure}

We first perform a validation check on the effectiveness of the proposed solver module under a non-learning setup, where the solver serves as a post-processing tool for depth refinement and completion. We start with a synthesis case where four different 100x100 planar regions are projected onto a 200x200 image. For the initial input, the geometry of each pixel has 95\% probability to be substituted with random noise, simulating unreliable pixels. Figure \ref{fig::toy} shows qualitative results of applying the iterative solver on the input geometry. With the iterative optimization of depths and normals, the geometry is gradually refined to be close to the groundtruth target, demonstrating that our proposed solver successfully integrate locally planar priors into the optimization. We further test the solver on real outdoor data acquired from AdelaideRMF dataset \cite{AdelaideRMF}, which is widely used for multi-structure fitting methods \cite{hee2013deterministic,magri2014t}. Specifically, we apply the iterative solver over the depth map acquired from the sparse reconstruction of COLMAP \cite{schonberger2016structure,schonberger2016pixelwise}. Results are shown in Figure \ref{fig::colmap}. Our proposed solver successfully complete the sparse input with reasonably correct output, demonstrating the effectiveness of the module. 

\subsection{Experiments on Multi-view Stereo}
\begin{figure*}[tb]
\scriptsize
\setlength\tabcolsep{1.0pt} 
\renewcommand{\arraystretch}{1.0}
\begin{tabular}{cccccccc}
{\includegraphics[width=0.11\linewidth]{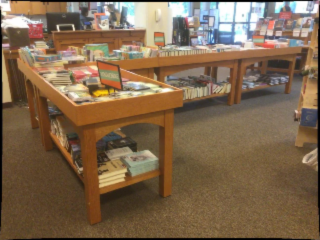}} & 
{\includegraphics[width=0.11\linewidth]{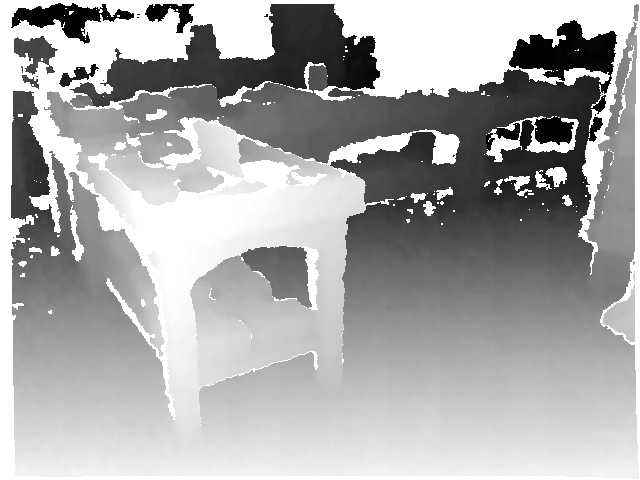}} &
{\includegraphics[width=0.11\linewidth]{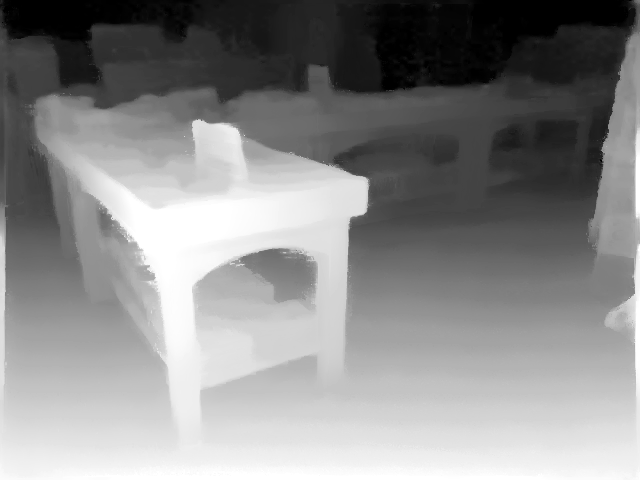}} & 
{\includegraphics[width=0.11\linewidth]{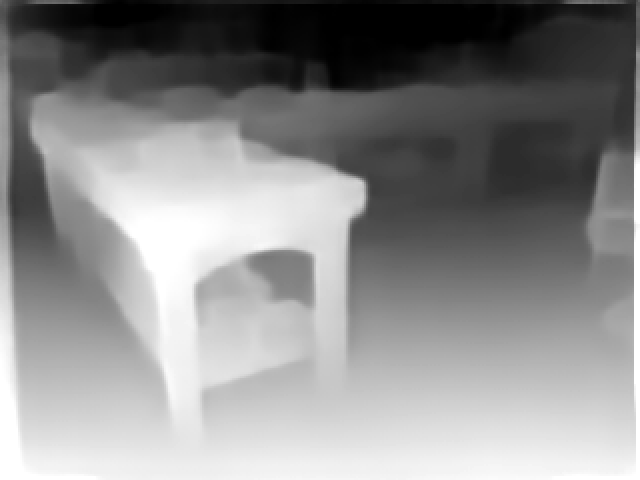}} & 
{\includegraphics[width=0.11\linewidth]{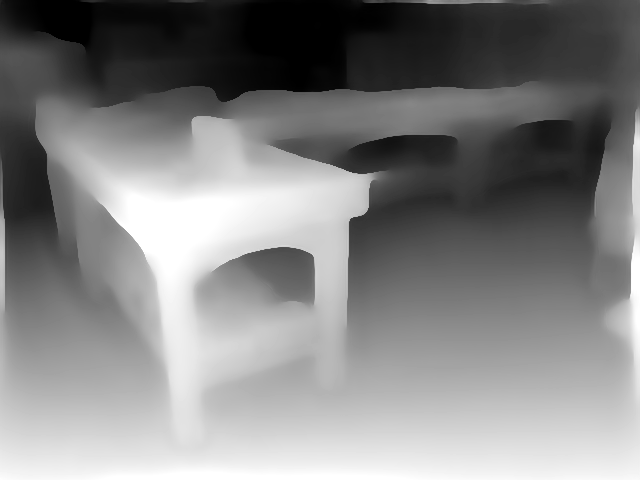}} &
{\includegraphics[width=0.11\linewidth]{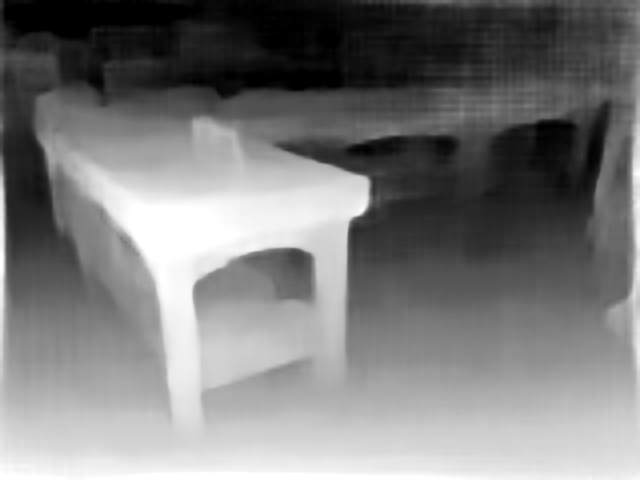}} &
{\includegraphics[width=0.11\linewidth]{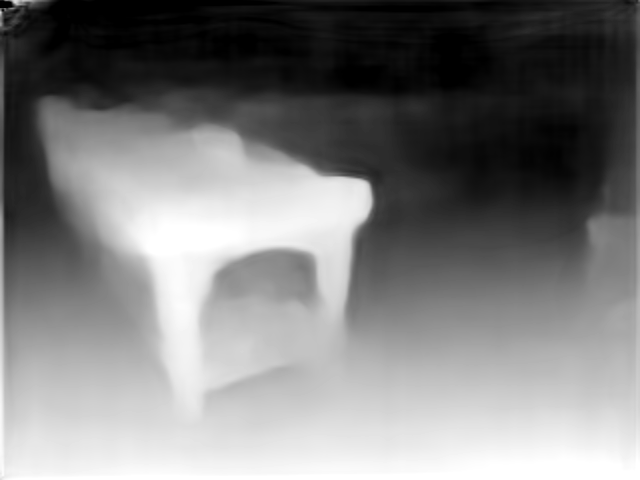}} &
{\includegraphics[width=0.11\linewidth]{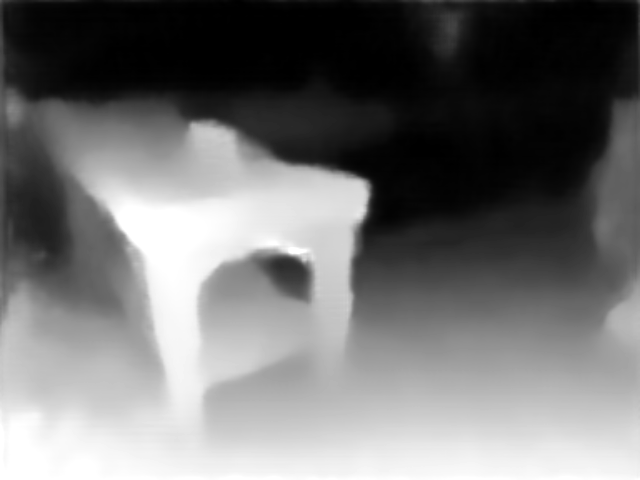}} \\



{\includegraphics[width=0.11\linewidth]{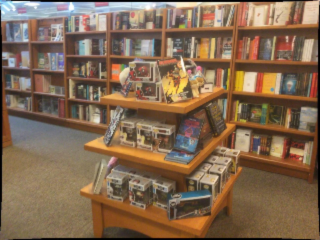}} & 
{\includegraphics[width=0.11\linewidth]{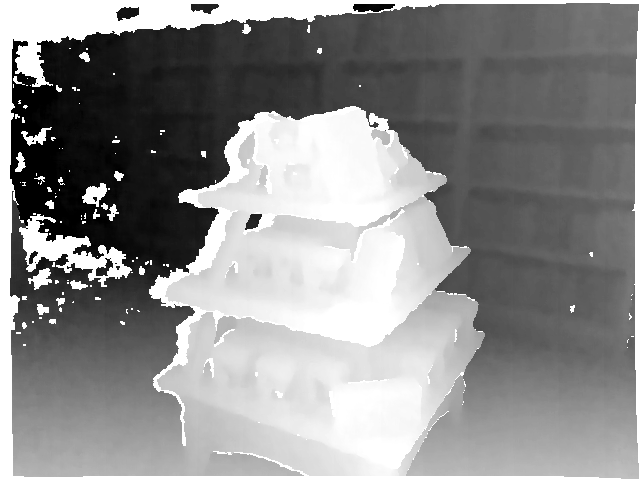}} &
{\includegraphics[width=0.11\linewidth]{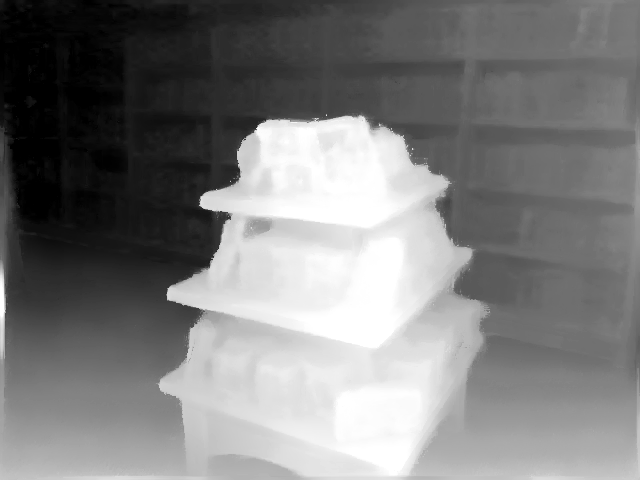}} & 
{\includegraphics[width=0.11\linewidth]{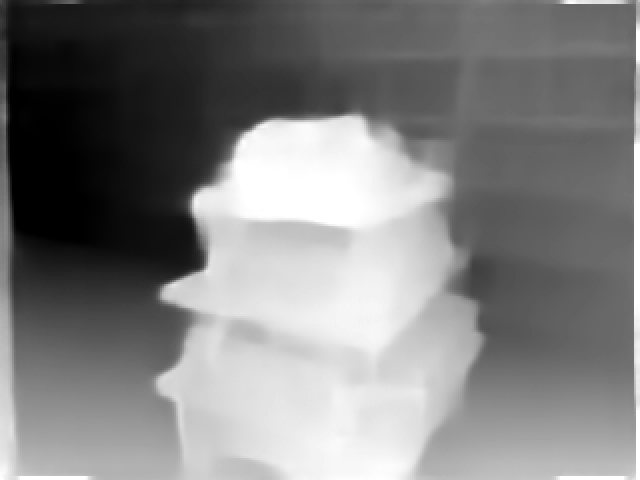}} & 
{\includegraphics[width=0.11\linewidth]{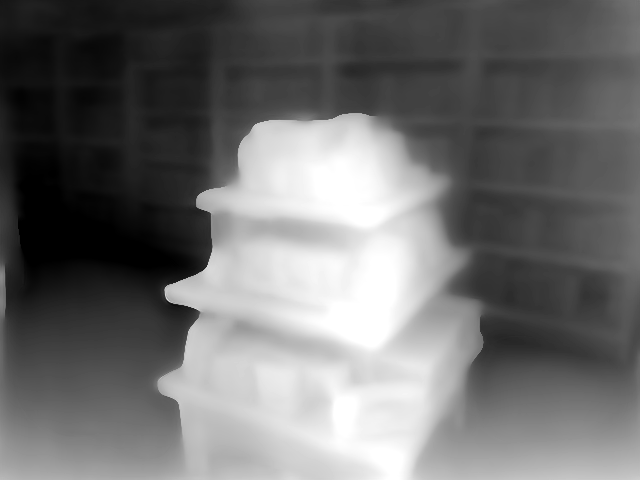}} &
{\includegraphics[width=0.11\linewidth]{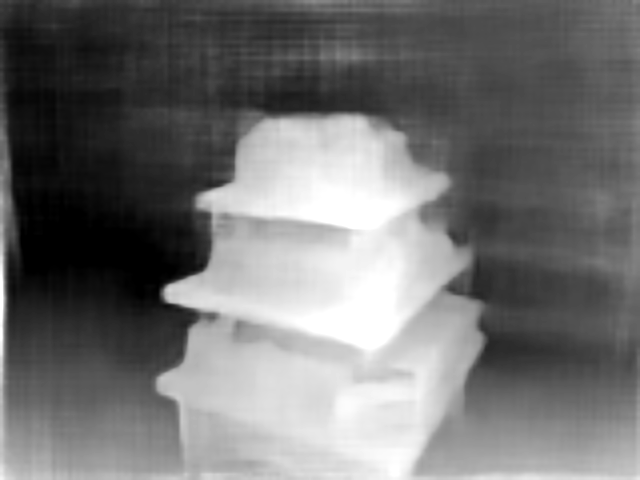}} &
{\includegraphics[width=0.11\linewidth]{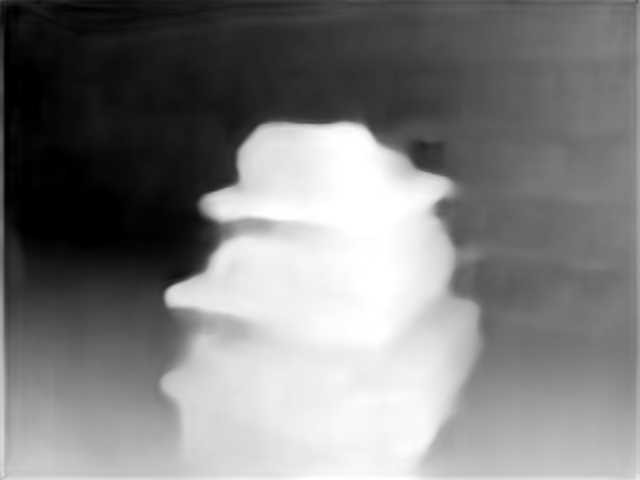}} &
{\includegraphics[width=0.11\linewidth]{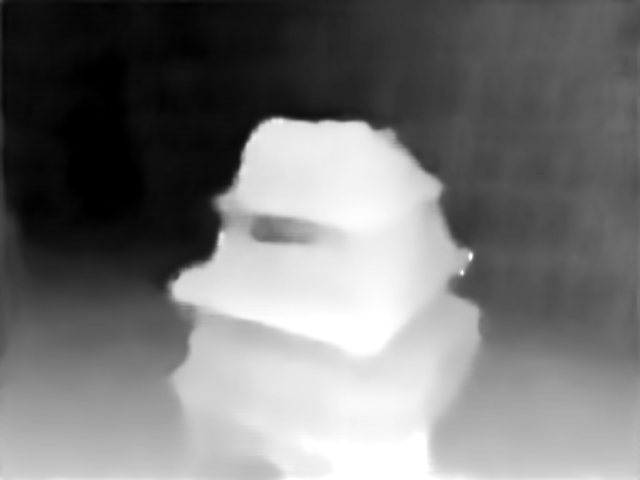}} \\

{\includegraphics[width=0.11\linewidth]{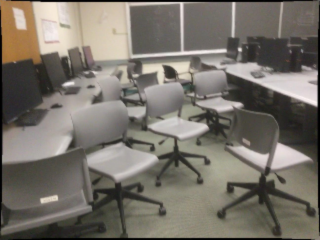}} & 
{\includegraphics[width=0.11\linewidth]{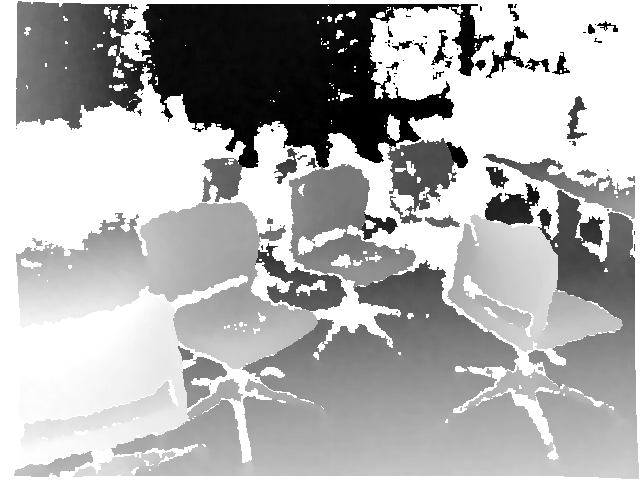}} &
{\includegraphics[width=0.11\linewidth]{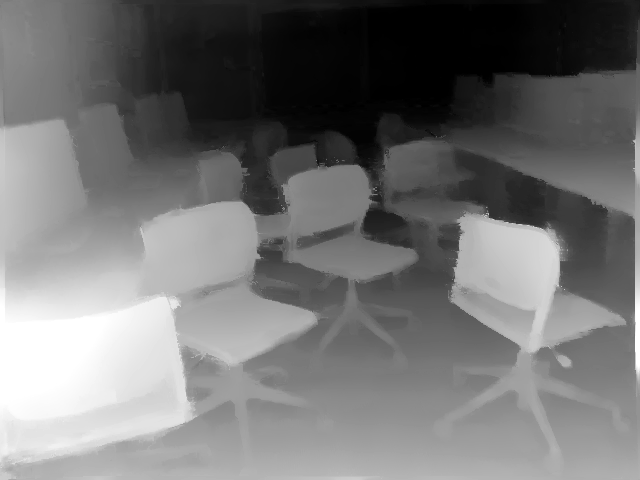}} & 
{\includegraphics[width=0.11\linewidth]{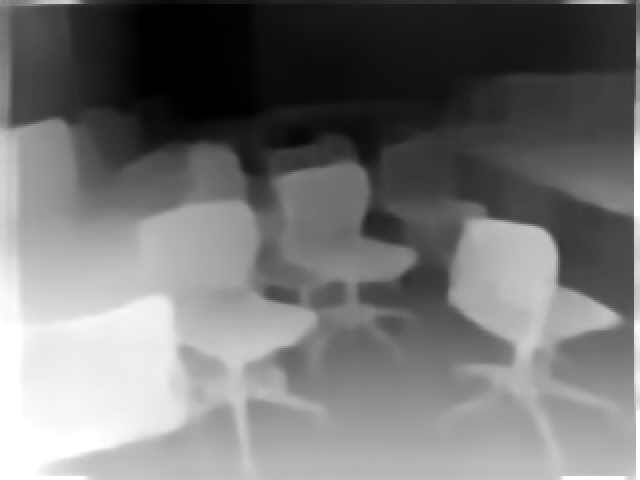}} & 
{\includegraphics[width=0.11\linewidth]{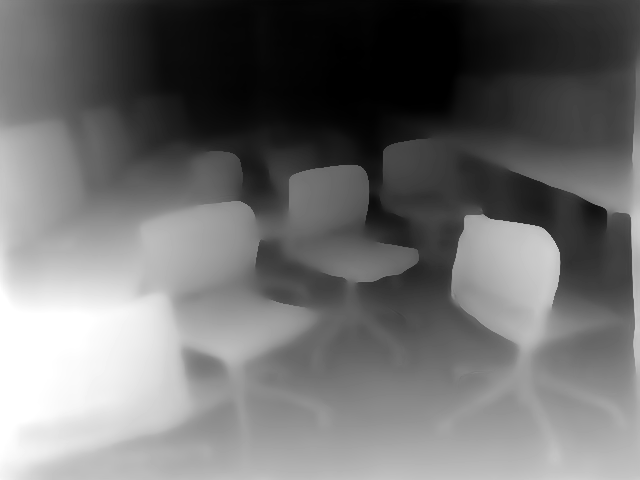}} &
{\includegraphics[width=0.11\linewidth]{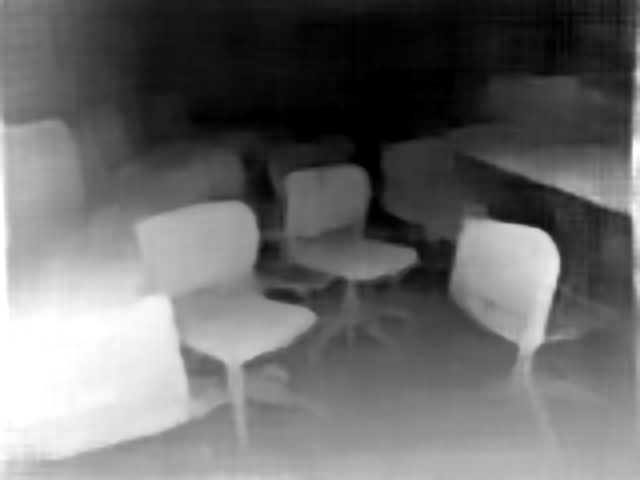}} &
{\includegraphics[width=0.11\linewidth]{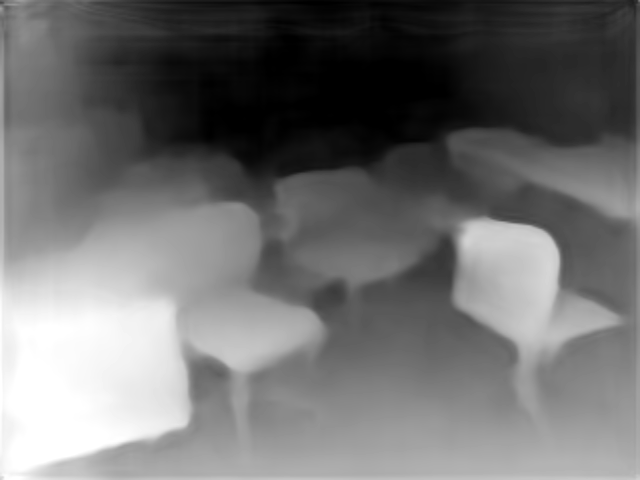}} &
{\includegraphics[width=0.11\linewidth]{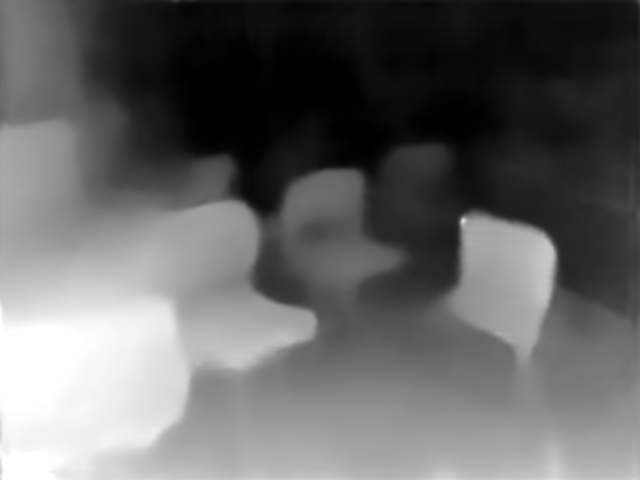}} \\

{\includegraphics[width=0.11\linewidth]{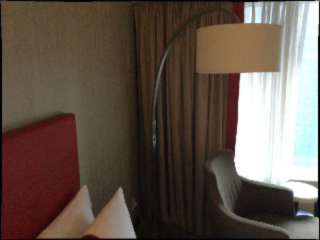}} & 
{\includegraphics[width=0.11\linewidth]{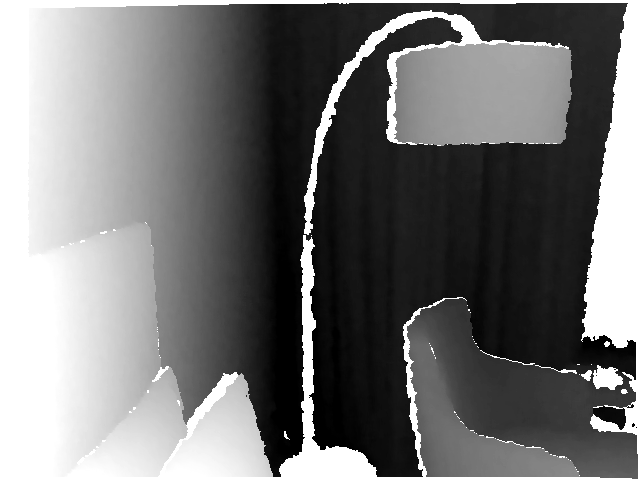}} &
{\includegraphics[width=0.11\linewidth]{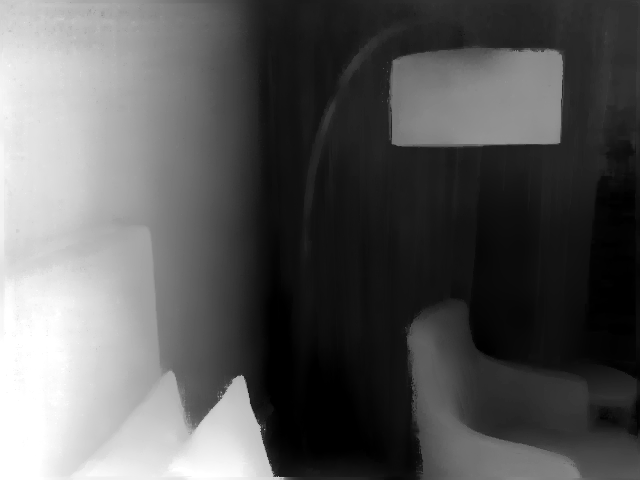}} & 
{\includegraphics[width=0.11\linewidth]{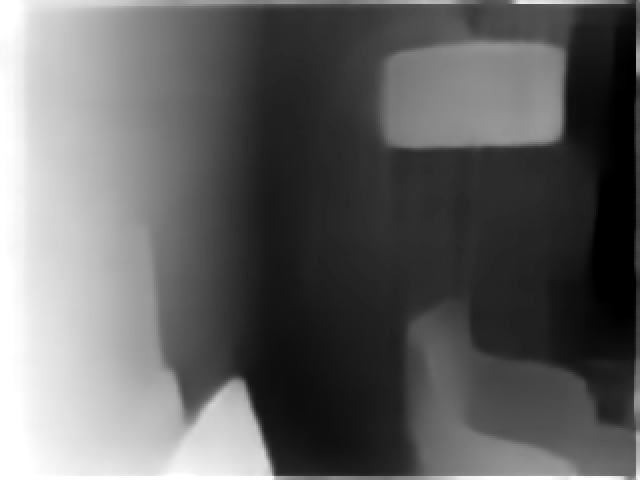}} & 
{\includegraphics[width=0.11\linewidth]{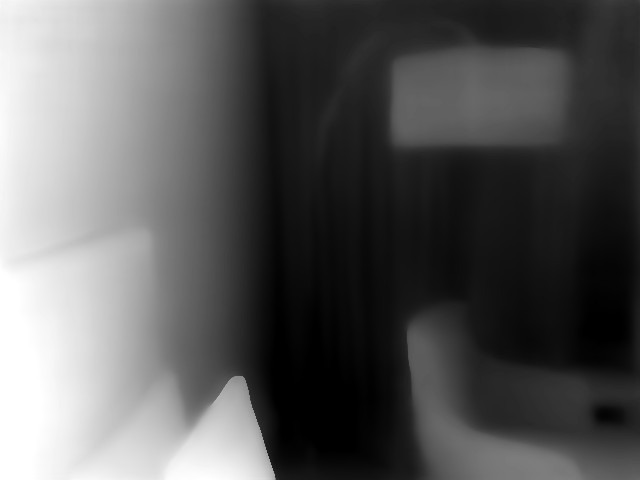}} &
{\includegraphics[width=0.11\linewidth]{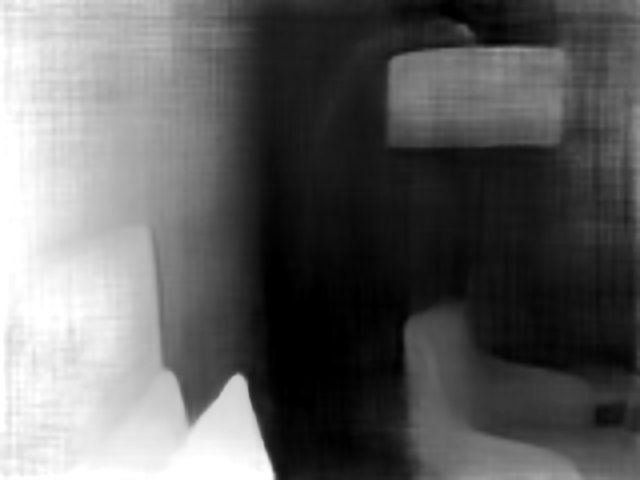}} &
{\includegraphics[width=0.11\linewidth]{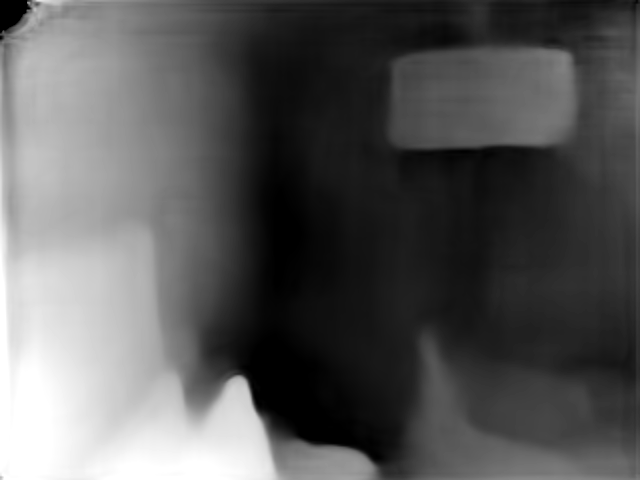}} &
{\includegraphics[width=0.11\linewidth]{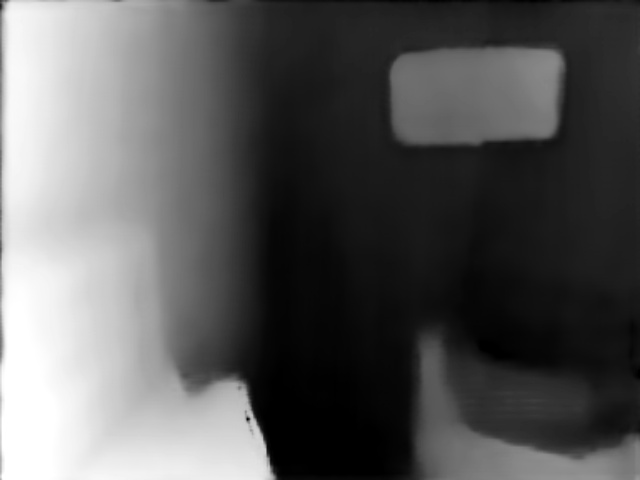}} \\

{\includegraphics[width=0.11\linewidth]{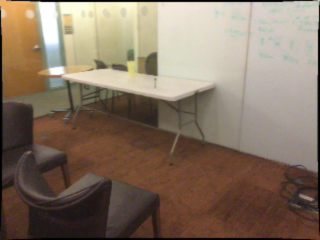}} & 
{\includegraphics[width=0.11\linewidth]{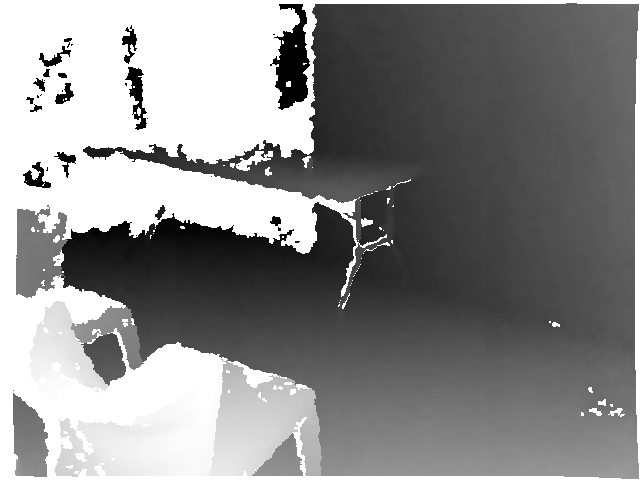}} &
{\includegraphics[width=0.11\linewidth]{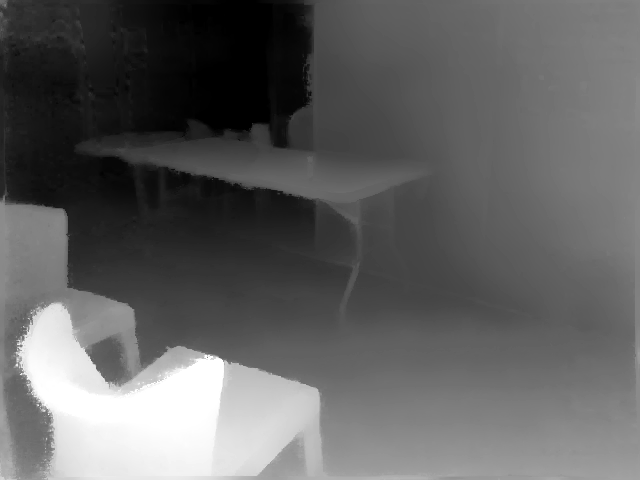}} & 
{\includegraphics[width=0.11\linewidth]{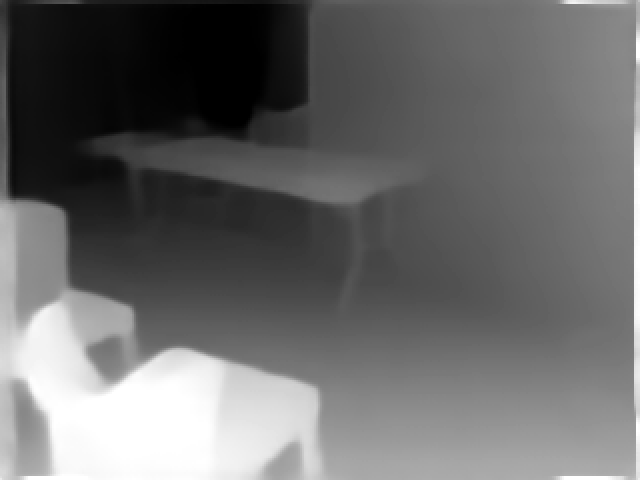}} & 
{\includegraphics[width=0.11\linewidth]{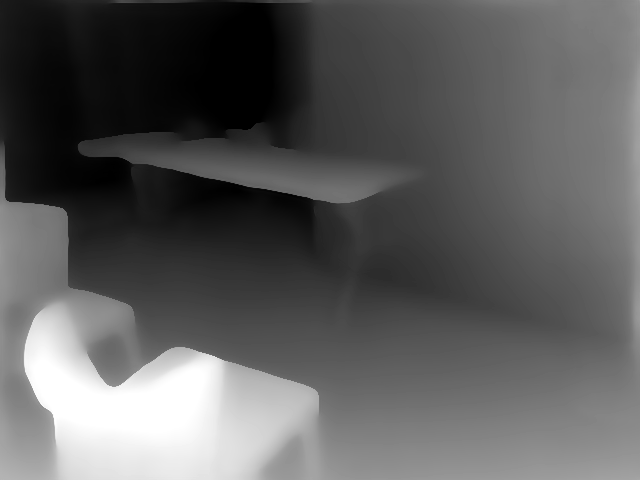}} &
{\includegraphics[width=0.11\linewidth]{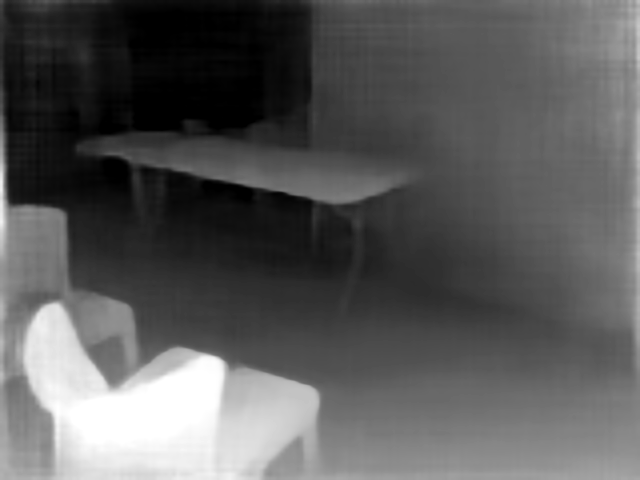}} &
{\includegraphics[width=0.11\linewidth]{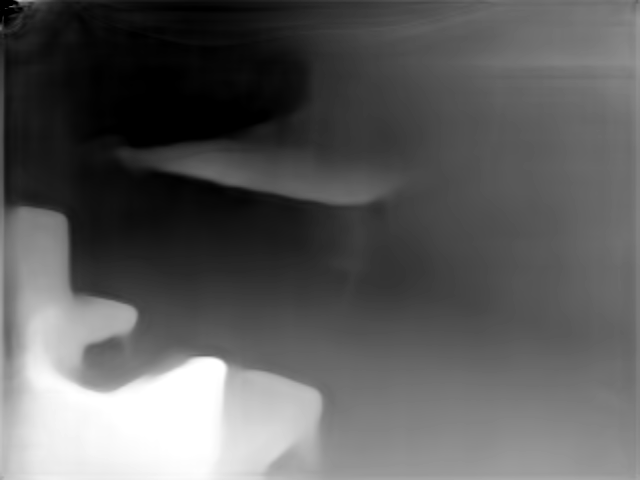}} &
{\includegraphics[width=0.11\linewidth]{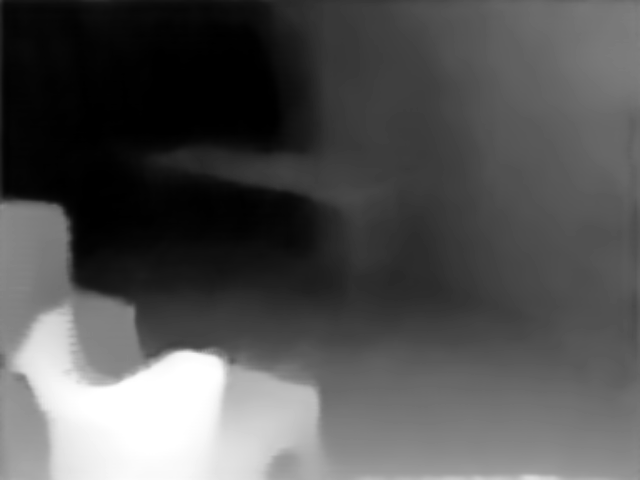}} \\

{\includegraphics[width=0.11\linewidth]{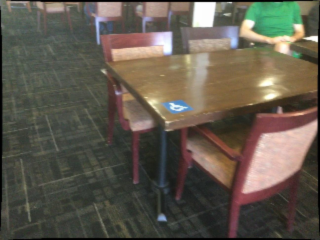}} & 
{\includegraphics[width=0.11\linewidth]{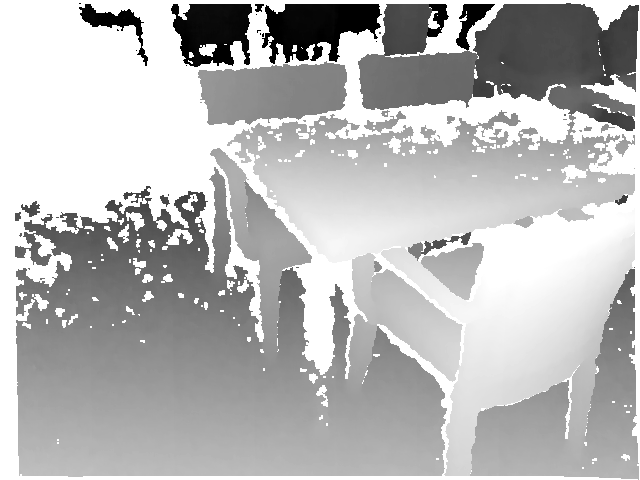}} &
{\includegraphics[width=0.11\linewidth]{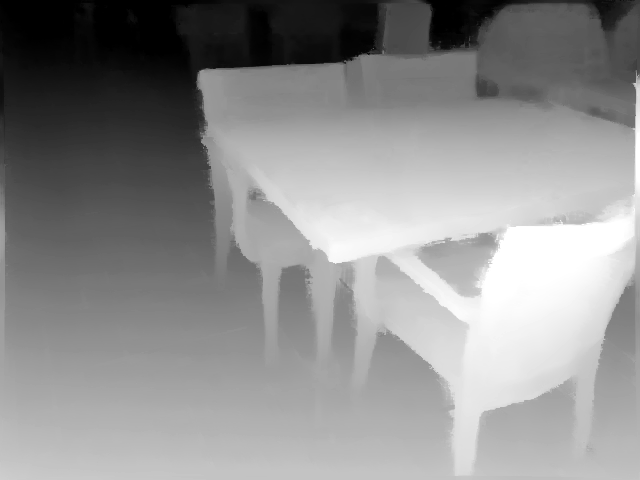}} & 
{\includegraphics[width=0.11\linewidth]{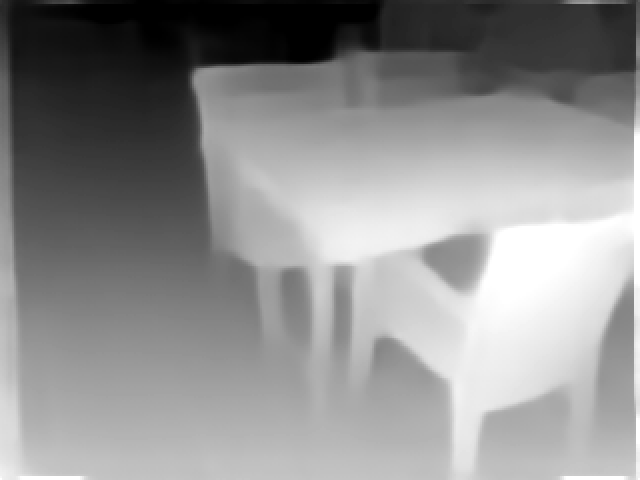}} & 
{\includegraphics[width=0.11\linewidth]{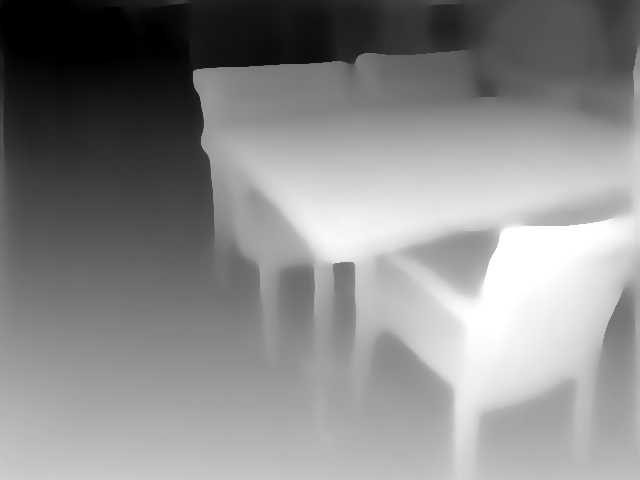}} &
{\includegraphics[width=0.11\linewidth]{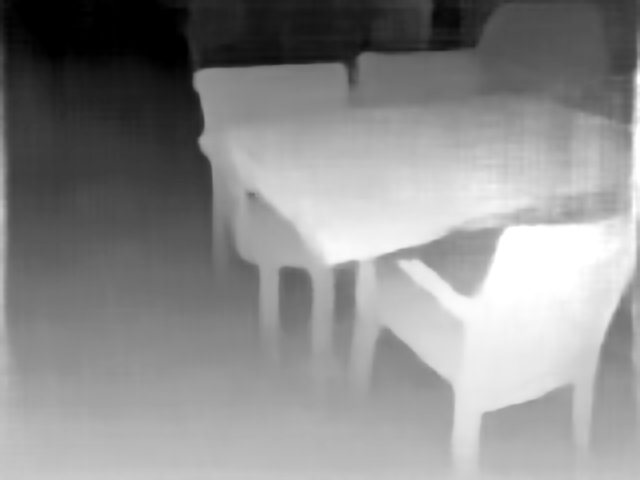}} &
{\includegraphics[width=0.11\linewidth]{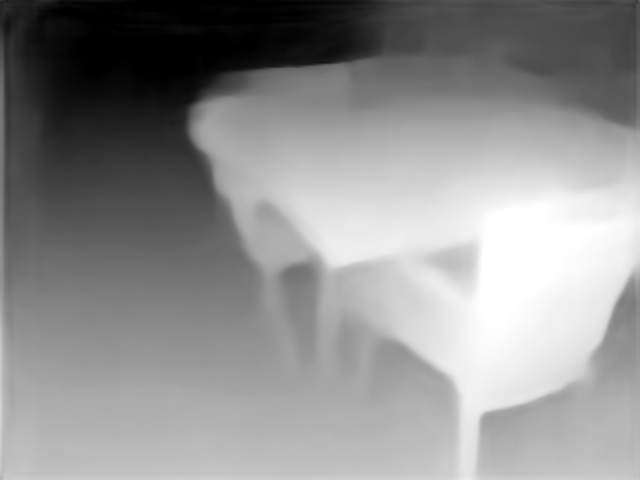}} &
{\includegraphics[width=0.11\linewidth]{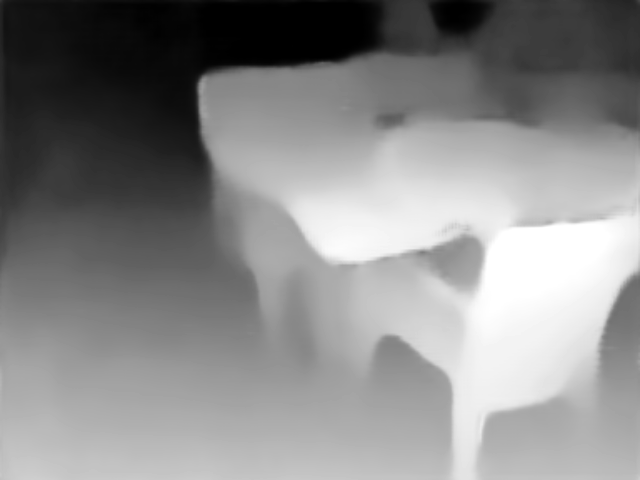}} \\

(a) Image & (b) Groundtruth & (c) Ours & (d) DELTAS \cite{sinha2020deltas} & (e) NAS \cite{kusupati2020normal} & (f) DPSNet (FT) \cite{im2019dpsnet} &  (g) MVDepth (FT) \cite{wang2018mvdepthnet} & (h) N-RGBD \cite{liu2019neural} \\

\end{tabular}
\vspace{0.05cm}
\centering
\caption{Qualititative results of multi-view depth estimation on ScanNet \cite{dai2017scannet}. \textbf{Better viewed when zoomed in.}}
\label{fig::depth_scannet}
\vspace{-10pt}
\end{figure*}

\begin{figure*}[tb]
\scriptsize
\setlength\tabcolsep{1.0pt} 
\renewcommand{\arraystretch}{1.0}
\begin{tabular}{cccc}
{\includegraphics[width=0.25\linewidth]{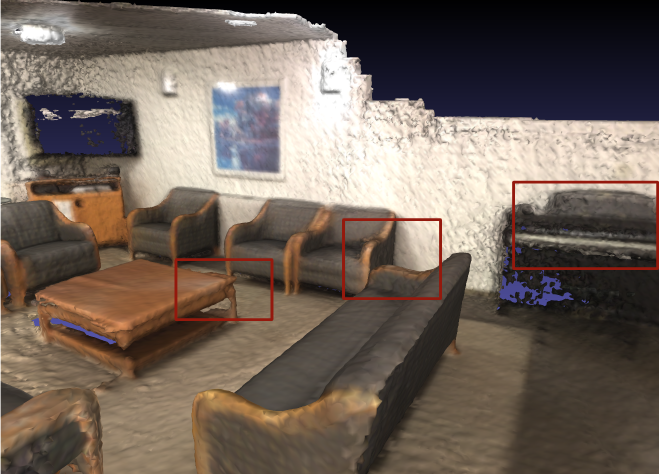}} & 
{\includegraphics[width=0.25\linewidth]{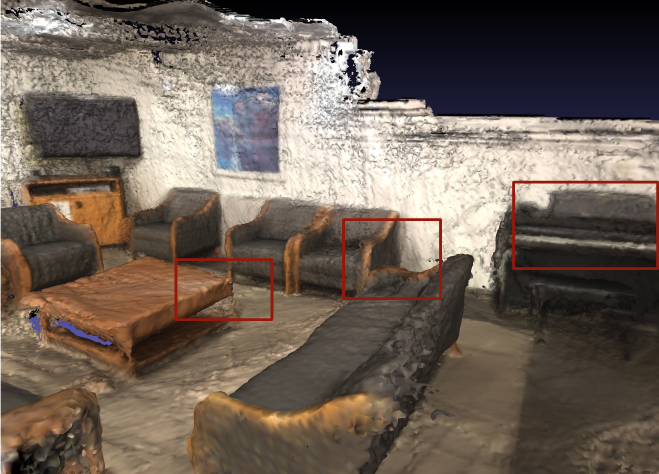}} &
{\includegraphics[width=0.25\linewidth]{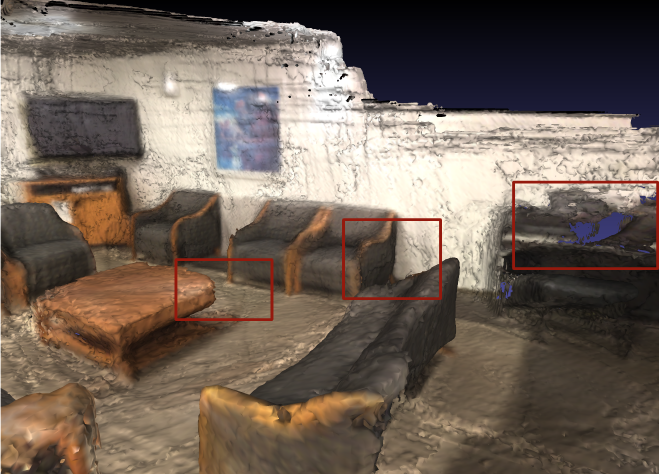}} & 
{\includegraphics[width=0.25\linewidth]{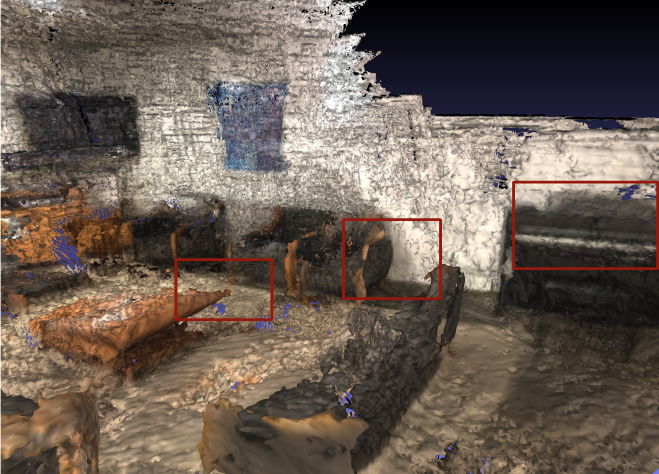}}  \\

(a) Groundtruth & (b) Ours & (c) DELTAS \cite{sinha2020deltas} & (d) DPS \cite{im2019dpsnet} \\

\end{tabular}
\centering
\vspace{0.05cm}
\caption{Visualization of our reconstruction results compared to DELTAS \cite{sinha2020deltas} and DPS \cite{im2019dpsnet}. \textbf{Better viewed when zoomed in.}}
\label{fig::reconstruction}
\vspace{-10pt}
\end{figure*}

\begin{figure}[tb]
\scriptsize
\setlength\tabcolsep{1.0pt} 
\begin{tabular}{ccccc}
Image & GT & CNM \cite{long2020occlusion} & NAS \cite{kusupati2020normal} & Ours \\
{\includegraphics[width=0.19\linewidth]{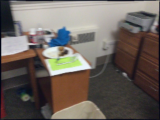}} & 
{\includegraphics[width=0.19\linewidth]{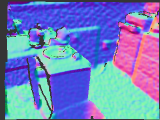}} &
{\includegraphics[width=0.19\linewidth]{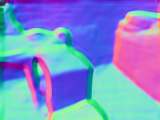}} &
{\includegraphics[width=0.19\linewidth]{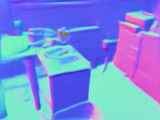}} &
{\includegraphics[width=0.19\linewidth]{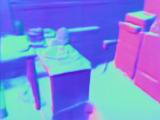}}
\\
{\includegraphics[width=0.19\linewidth]{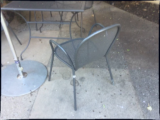}} & 
{\includegraphics[width=0.19\linewidth]{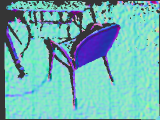}} &
{\includegraphics[width=0.19\linewidth]{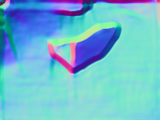}} &
{\includegraphics[width=0.19\linewidth]{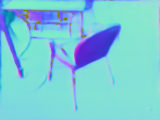}} &
{\includegraphics[width=0.19\linewidth]{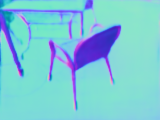}}
\\
{\includegraphics[width=0.19\linewidth]{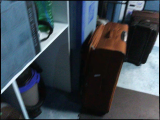}} & 
{\includegraphics[width=0.19\linewidth]{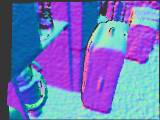}} &
{\includegraphics[width=0.19\linewidth]{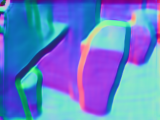}} &
{\includegraphics[width=0.19\linewidth]{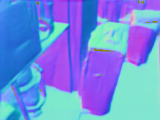}} &
{\includegraphics[width=0.19\linewidth]{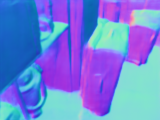}}
\end{tabular}
\centering
\vspace{0.05cm}
\caption{Qualitative results of surface normal estimation on ScanNet \cite{dai2017scannet}. Refer to supp. for higher-resolution images.}
\label{fig::normal_scannet}
\vspace{-20pt}
\end{figure}

To compare our proposed deep MVS system with leading approaches, we evaluate our method on well-established benchmarks including ScanNet's official test split \cite{dai2017scannet} and RGB-D Scenes V2 dataset \cite{lai2014unsupervised}. Both datasets include challenging indoor scenes with many poorly textured regions. Quantitative results on multi-view depth estimation are shown in Table \ref{tab::depth-scannet} and Table \ref{tab::depth-rgbd}, where our system surpasses all existing state-of-the-art methods by a large margin on both datasets. Qualitative results in Figure \ref{fig::depth_scannet} also demonstrate great improvement of our proposed system. 
Our method not only produces accurate depth map, but also successfully handles fine details around boundaries. 
We further evaluate the estimated surface normal for our deep MVS system. Results are shown in Table \ref{tab::normal-scannet}, where we again achieves state-of-the-art performance on ScanNet dataset \cite{dai2017scannet}. Qualitative visualizations in Figure \ref{fig::normal_scannet} clearly show the improvement of our method. Finally, we show visualizations of reconstructed models after TSDF fusion \cite{curless1996volumetric} in Figure \ref{fig::reconstruction}. Our method produces visually more appealing reconstructions compared to strong baseline methods \cite{im2019dpsnet,sinha2020deltas}.
\begin{table*}[tb]
\begin{center}
\begin{tabular}{l|ccccc|ccc}
\hline
Method & Abs Rel & Abs Diff & Sq Rel & RMSE & RMSE log & $\delta < 1.25$ & $\delta < 1.25^2$ & $\delta < 1.25^3$ \\
\hline
MVDepth \cite{wang2018mvdepthnet} & 0.1053 & 0.1987 & 0.0634 & 0.3026 & 0.1490 & 0.8817 & 0.9723 & 0.9924 \\
MVDepth (FT) & 0.1014 & 0.1891 & 0.0476 & 0.2850 & 0.1390 & 0.8930 & 0.9764 & 0.9941 \\
GP-MVS \cite{hou2019multi} & 0.0920 & 0.2283 & 0.0644 & 0.4436 & 0.1560 & 0.8918 & 0.9629 & 0.9918 \\
GP-MVS (FT) & 0.0787 & 0.2008 & 0.0518 & 0.4009 & 0.1394 & 0.9134 & 0.9643 & 0.9931 \\ 
NeuralRGBD \cite{liu2019neural} & 0.0871 & 0.1710 & 0.0409 & 0.2693 & 0.1324 & 0.9150 & 0.9785 & 0.9925 \\
CNM \cite{long2020occlusion} & 0.1119 & 0.2101 & 0.0510 & 0.2970 & 0.1485 & 0.8686 & 0.9724 & 0.9930 \\
DPSNet \cite{im2019dpsnet} & 0.1164 & 0.1992 & 0.0606 & 0.3065 & 0.1602 & 0.8569 & 0.9575 & 0.9884 \\
DPSNet (FT) & 0.0910 & 0.1807 & 0.0410 & 0.2697 & 0.1291 & 0.9008 & 0.9787 & 0.9952 \\
NAS \cite{kusupati2020normal} & 0.0795 & 0.1597 & 0.0323 & 0.2357 & 0.1112 & 0.9284 & 0.9862 & 0.9966 \\
DELTAS \cite{sinha2020deltas} & 0.0738 & 0.1380 & 0.0245 & 0.2051 & 0.1021 & 0.9473 & 0.9890 & 0.9976 \\
Ours & \textbf{0.0665} & \textbf{0.1281} & \textbf{0.0240} & \textbf{0.1995} & \textbf{0.0990} & \textbf{0.9489} & \textbf{0.9896} & \textbf{0.9978} \\
\hline
\end{tabular}
\vspace{0.1cm}
\caption{Quantitative comparisons between our method and state-of-the-art deep MVS methods on ScanNet dataset \cite{dai2017scannet}. All the methods use sequences of length 3 and fixed reference interval 20 for testing, except GP-MVS \cite{hou2019multi} and NeuralRGBD \cite{liu2019neural}, which directly use the whole sequence. Since \cite{wang2018mvdepthnet, hou2019multi,im2019dpsnet} were not initially trained on ScanNet, we also report the results after finetuning, denoted as ``FT".}
\label{tab::depth-scannet}
\end{center}
\vspace{-10pt}
\end{table*}
\begin{table}[tb]
\begin{center}
\begin{tabular}{l|cc|ccc}
\hline
Method & Mean & Median & 11.25\textdegree & 22.5\textdegree & 30\textdegree \\
CNM \cite{long2020occlusion} & 27.92 & 22.12 & 27.43 & 52.16 & 63.44 \\
NAS \cite{kusupati2020normal} & 24.12 & 18.02 & 31.59 & 60.20 & 69.45 \\
Ours & \textbf{22.30} & \textbf{16.75} & \textbf{34.80} & \textbf{64.39} & \textbf{75.11} \\
\hline
\end{tabular}
\vspace{0.1cm}
\caption{Quantitative comparisons of surface normal estimation between our method and state-of-the-art methods \cite{long2020occlusion,kusupati2020normal}.}
\label{tab::normal-scannet}
\end{center}
\vspace{-10pt}
\end{table}

\begin{table}[tb]
\begin{center}
\scalebox{0.8} {
\begin{tabular}{l|ccccc}
\hline
Method & Abs Rel & Abs Diff & Sq Rel & RMSE & $\delta < 1.25$ \\
\hline
MVDepth \cite{wang2018mvdepthnet} & 0.0885 & 0.1467 & 0.0314 & 0.2313 & 0.9184 \\
GP-MVS \cite{hou2019multi} & 0.1087 & 0.1514 & 0.0827 & 0.2873 & 0.9170 \\
N-RGBD \cite{liu2019neural} & 0.0995 & 0.1530 & 0.0352 & 0.2361 & 0.9233 \\
CNM \cite{long2020occlusion} & 0.1350 & 0.1873 & 0.0484 & 0.2619 & 0.8667\\
DPSNet \cite{im2019dpsnet} & 0.0771 & 0.1290 & 0.0234 & 0.2045 & 0.9401\\
NAS \cite{kusupati2020normal} & 0.0732 & 0.1241 & 0.0198 & 0.1893 & 0.9576 \\ 
DELTAS \cite{sinha2020deltas} & 0.1065 & 0.1528 & 0.0299 & 0.2138 & 0.9156 \\
Ours & \textbf{0.0698} & \textbf{0.1130} & \textbf{0.0194} & \textbf{0.1770} & \textbf{0.9681} \\
\hline
\end{tabular}
}
\vspace{0.1cm}
\caption{Quantitative comparisons between our method and state-of-the-art deep MVS methods on RGB-D Scenes V2 dataset \cite{lai2014unsupervised}.}
\label{tab::depth-rgbd}
\end{center}
\vspace{-20pt}
\end{table}



\subsection{Ablation Studies}
We perform several ablation studies to further understand the behaviors of the proposed solver. All ablation studies are conducted on ScanNet \cite{dai2017scannet}. To study the contribution of the depth-normal solver with its post processing and end-to-end joint training, we train a baseline with the same network architecture completely without the proposed solver, which does not benefit from joint training. For comparison, we also apply the solver to refine the baseline's output with the hybrid confidence map at inference. Three conclusions can be acquired from the results shown in Table \ref{tab::ablation-filter}: 1) As a post-processing tool, the proposed solver consistently improves the output of the network no matter whether the network is jointly trained. 2) The joint training improves the quality of the estimated initial depth (0.0735 $\rightarrow$ 0.0711). This is because joint training gets the network to focus more on reliable estimation on well textured areas, which largely releases the learning burden. 3) With end-to-end joint training, the performance gain of applying the solver module at inference increases. This agrees with our assumption that during end-to-end training, the network is guided to output initial geometry that well fits the subsequent solver module. 

In addition, we also study the design choices on the confidence map. Table \ref{tab::ablation-conf} shows the results. We observe that geometric confidence is more effective than the estimated deep confidence. Although the performance gain of including the deep confidence is not significant, the deep confidence is complete and will not be affected by occlusions or small overlaps, which can serve as a good supplements for geometric confidence in practice. Furthermore, we study the effects of different number of reference views during geometric consistency check and different number of iterations at inference. Table \ref{tab::ablation-iter-view} shows the results, which indicate that improving the number of iterations consistently improve the final depth quality. However, as currently we take the minimum confidence over all reference frames to get a ``strict" depth confidence, adding views will only result in slight performance gain. More advanced confidence based on multi-view depth fusion is left for future work.

\begin{table}[tb]
\begin{center}
\begin{tabular}{l|cc|c}
\hline
Method & Abs Rel & RMSE & $\delta<1.25$ \\
\hline
Baseline & 0.0735 & 0.2169 & 0.9359\\
Baseline + post & \textbf{0.0720} & \textbf{0.2148} & \textbf{0.9378}\\
\hline
Joint train w/o post & 0.0711 & 0.2121 & 0.9384\\
Joint train + post & \textbf{0.0665} & \textbf{0.1995} & \textbf{0.9489}\\
\hline
\end{tabular}
\vspace{0.1cm}
\caption{Ablation studies on the contribution of the solver used at post-processing and end-to-end joint training. ``post" indicates using the solver to post-process the depth map with 10 iterations.}
\label{tab::ablation-filter}
\end{center}
\vspace{-10pt}
\end{table}



\begin{table}[tb]
\begin{center}
\begin{tabular}{l|cc|c}
\hline
Method & Abs Rel & RMSE & $\delta<1.25$\\
\hline
Joint train w/o post & 0.0711 & 0.2121 & 0.9384 \\
Groundtruth conf & 0.0599 & 0.1931 & 0.9604\\
\hline
Deep conf only & 0.0692 & 0.2040 & 0.9421\\
Geometric conf only & 0.0667 & 0.2001 & 0.9486\\
Deep + geometric conf & \textbf{0.0665} & \textbf{0.1995} & \textbf{0.9489} \\
\hline
\end{tabular}
\vspace{0.1cm}
\caption{Studies on the choices of different confidence maps.}
\label{tab::ablation-conf}
\end{center}
\vspace{-10pt}
\end{table}


\begin{table}[tb]
\begin{center}
\begin{tabular}{l|cccc}
\hline
View/Iter & 1 iter & 5 iters & 10 iters & 25 iters \\
\hline
1 view & 0.0694 & 0.0680 & 0.0675 & 0.0670 \\ 
2 views & 0.0690 & 0.0672 & 0.0665 & 0.0659 \\
4 views & 0.0688 & 0.0669 & 0.0662 & 0.0655 \\
\hline
\end{tabular}
\vspace{0.1cm}
\caption{Ablation studies on the effects of different number of views used during geometric consistency check and different number of iterations at inference. Abs Rel is reported for each entry.}
\label{tab::ablation-iter-view}
\end{center}
\vspace{-20pt}
\end{table}



\section{Conclusion}
In this work, we present a deep MVS system with a novel confidence-based iterative depth-normal solver. We formulate an energy potential that favors locally planar surfaces and propose to perform iterative subproblem optimization over depths and surface normals in turn. Not only is the solver able to serve as a post-processing tool for plane-based depth refinement and completion, but it is also in closed form such that it can be integrated into our deep MVS system with end-to-end joint training. 
Future directions include advanced confidence estimation and spatial-temporal plane-based propagation.

\noindent
\textbf{Acknowledgements:}
This work was partially supported by the Natural Science Foundation of China (61725204),  BNRist and Tsinghua University (CS Dept) - DeepBlue Technology (Shanghai) Company Limited Joint Research Center for Machine Vision (JCMV).

\clearpage
{\small
\bibliographystyle{ieee_fullname}
\bibliography{egbib}

\begin{thebibliography}{10}\itemsep=-1pt

\bibitem{AdelaideRMF}
http://cs.adelaide.edu.au/~hwong/doku.php?id=data.

\bibitem{barron2015fast}
Jonathan~T Barron, Andrew Adams, YiChang Shih, and Carlos Hern{\'a}ndez.
\newblock Fast bilateral-space stereo for synthetic defocus.
\newblock In {\em CVPR}, pages 4466--4474, 2015.

\bibitem{barron2016fast}
Jonathan~T Barron and Ben Poole.
\newblock The fast bilateral solver.
\newblock In {\em ECCV}, pages 617--632. Springer, 2016.

\bibitem{birchfield1999multiway}
Stan Birchfield and Carlo Tomasi.
\newblock Multiway cut for stereo and motion with slanted surfaces.
\newblock In {\em ICCV}, volume~1, pages 489--495, 1999.

\bibitem{bleyer2005layered}
Michael Bleyer and Margrit Gelautz.
\newblock A layered stereo matching algorithm using image segmentation and
  global visibility constraints.
\newblock {\em ISPRS Journal of Photogrammetry and remote sensing},
  59(3):128--150, 2005.

\bibitem{bleyer2011patchmatch}
Michael Bleyer, Christoph Rhemann, and Carsten Rother.
\newblock Patchmatch stereo-stereo matching with slanted support windows.
\newblock In {\em BMVC}, volume~11, pages 1--11, 2011.

\bibitem{chen2019point}
Rui Chen, Songfang Han, Jing Xu, and Hao Su.
\newblock Point-based multi-view stereo network.
\newblock In {\em ICCV}, pages 1538--1547, 2019.

\bibitem{curless1996volumetric}
Brian Curless and Marc Levoy.
\newblock A volumetric method for building complex models from range images.
\newblock In {\em Proceedings of the 23rd annual conference on Computer
  graphics and interactive techniques}, pages 303--312, 1996.

\bibitem{dai2017scannet}
Angela Dai, Angel~X Chang, Manolis Savva, Maciej Halber, Thomas Funkhouser, and
  Matthias Nie{\ss}ner.
\newblock Scannet: Richly-annotated 3d reconstructions of indoor scenes.
\newblock In {\em CVPR}, pages 5828--5839, 2017.

\bibitem{duzcceker2020deepvideomvs}
Arda D{\"u}z{\c{c}}eker, Silvano Galliani, Christoph Vogel, Pablo Speciale,
  Mihai Dusmanu, and Marc Pollefeys.
\newblock Deepvideomvs: Multi-view stereo on video with recurrent
  spatio-temporal fusion.
\newblock In {\em CVPR}, 2021.

\bibitem{eigen2015predicting}
David Eigen and Rob Fergus.
\newblock Predicting depth, surface normals and semantic labels with a common
  multi-scale convolutional architecture.
\newblock In {\em ICCV}, pages 2650--2658, 2015.

\bibitem{faugeras2002variational}
Olivier Faugeras and Renaud Keriven.
\newblock {\em Variational principles, surface evolution, PDE's, level set
  methods and the stereo problem}.
\newblock IEEE, 2002.

\bibitem{galliani2015massively}
Silvano Galliani, Katrin Lasinger, and Konrad Schindler.
\newblock Massively parallel multiview stereopsis by surface normal diffusion.
\newblock In {\em ICCV}, pages 873--881, 2015.

\bibitem{gallup2007real}
David Gallup, Jan-Michael Frahm, Philippos Mordohai, Qingxiong Yang, and Marc
  Pollefeys.
\newblock Real-time plane-sweeping stereo with multiple sweeping directions.
\newblock In {\em CVPR}, pages 1--8, 2007.

\bibitem{graber2015efficient}
Gottfried Graber, Jonathan Balzer, Stefano Soatto, and Thomas Pock.
\newblock Efficient minimal-surface regularization of perspective depth maps in
  variational stereo.
\newblock In {\em CVPR}, pages 511--520, 2015.

\bibitem{hee2013deterministic}
Kwang Hee~Lee and Sang Wook~Lee.
\newblock Deterministic fitting of multiple structures using iterative maxfs
  with inlier scale estimation.
\newblock In {\em ICCV}, pages 41--48, 2013.

\bibitem{hirschmuller2007stereo}
Heiko Hirschmuller.
\newblock Stereo processing by semiglobal matching and mutual information.
\newblock {\em PAMI}, 30(2):328--341, 2007.

\bibitem{hirschmuller2008evaluation}
Heiko Hirschmuller and Daniel Scharstein.
\newblock Evaluation of stereo matching costs on images with radiometric
  differences.
\newblock {\em PAMI}, 31(9):1582--1599, 2008.

\bibitem{hong2004segment}
Li Hong and George Chen.
\newblock Segment-based stereo matching using graph cuts.
\newblock In {\em CVPR}, volume~1, pages I--I, 2004.

\bibitem{hosni2012fast}
Asmaa Hosni, Christoph Rhemann, Michael Bleyer, Carsten Rother, and Margrit
  Gelautz.
\newblock Fast cost-volume filtering for visual correspondence and beyond.
\newblock {\em PAMI}, 35(2):504--511, 2012.

\bibitem{hou2019multi}
Yuxin Hou, Juho Kannala, and Arno Solin.
\newblock Multi-view stereo by temporal nonparametric fusion.
\newblock In {\em ICCV}, pages 2651--2660, 2019.

\bibitem{hu2012quantitative}
Xiaoyan Hu and Philippos Mordohai.
\newblock A quantitative evaluation of confidence measures for stereo vision.
\newblock {\em PAMI}, 34(11):2121--2133, 2012.

\bibitem{im2019dpsnet}
Sunghoon Im, Hae-Gon Jeon, Stephen Lin, and In~So Kweon.
\newblock Dpsnet: End-to-end deep plane sweep stereo.
\newblock In {\em ICLR}, 2019.

\bibitem{ji2017surfacenet}
Mengqi Ji, Juergen Gall, Haitian Zheng, Yebin Liu, and Lu Fang.
\newblock Surfacenet: An end-to-end 3d neural network for multiview stereopsis.
\newblock In {\em ICCV}, pages 2307--2315, 2017.

\bibitem{kar2017learning}
Abhishek Kar, Christian H{\"a}ne, and Jitendra Malik.
\newblock Learning a multi-view stereo machine.
\newblock In {\em NeurIPS}, 2017.

\bibitem{kingma2014adam}
Diederik~P Kingma and Jimmy Ba.
\newblock Adam: A method for stochastic optimization.
\newblock {\em arXiv preprint arXiv:1412.6980}, 2014.

\bibitem{kuhn2019plane}
Andreas Kuhn, Shan Lin, and Oliver Erdler.
\newblock Plane completion and filtering for multi-view stereo reconstruction.
\newblock In {\em GCPR}, pages 18--32. Springer, 2019.

\bibitem{kuhn2020deepc}
Andreas Kuhn, Christian Sormann, Mattia Rossi, Oliver Erdler, and Friedrich
  Fraundorfer.
\newblock Deepc-mvs: Deep confidence prediction for multi-view stereo
  reconstruction.
\newblock In {\em 3DV}, pages 404--413. IEEE, 2020.

\bibitem{kusupati2020normal}
Uday Kusupati, Shuo Cheng, Rui Chen, and Hao Su.
\newblock Normal assisted stereo depth estimation.
\newblock In {\em CVPR}, pages 2189--2199, 2020.

\bibitem{kutulakos2000theory}
Kiriakos~N Kutulakos and Steven~M Seitz.
\newblock A theory of shape by space carving.
\newblock {\em IJCV}, 38(3):199--218, 2000.

\bibitem{lai2014unsupervised}
Kevin Lai, Liefeng Bo, and Dieter Fox.
\newblock Unsupervised feature learning for 3d scene labeling.
\newblock In {\em ICRA}, pages 3050--3057. IEEE, 2014.

\bibitem{liu2019neural}
Chao Liu, Jinwei Gu, Kihwan Kim, Srinivasa~G Narasimhan, and Jan Kautz.
\newblock Neural rgb (r) d sensing: Depth and uncertainty from a video camera.
\newblock In {\em CVPR}, pages 10986--10995, 2019.

\bibitem{long2020occlusion}
Xiaoxiao Long, Lingjie Liu, Christian Theobalt, and Wenping Wang.
\newblock Occlusion-aware depth estimation with adaptive normal constraints.
\newblock In {\em ECCV}, pages 640--657. Springer, 2020.

\bibitem{magri2014t}
Luca Magri and Andrea Fusiello.
\newblock T-linkage: A continuous relaxation of j-linkage for multi-model
  fitting.
\newblock In {\em CVPR}, pages 3954--3961, 2014.

\bibitem{murez2020atlas}
Zak Murez, Tarrence van As, James Bartolozzi, Ayan Sinha, Vijay Badrinarayanan,
  and Andrew Rabinovich.
\newblock Atlas: End-to-end 3d scene reconstruction from posed images.
\newblock In {\em ECCV}, 2020.

\bibitem{patil2020don}
Vaishakh Patil, Wouter Van~Gansbeke, Dengxin Dai, and Luc Van~Gool.
\newblock Don’t forget the past: Recurrent depth estimation from monocular
  video.
\newblock {\em IEEE Robotics and Automation Letters}, 5(4):6813--6820, 2020.

\bibitem{poggi2016learning}
Matteo Poggi and Stefano Mattoccia.
\newblock Learning from scratch a confidence measure.
\newblock In {\em BMVC}, 2016.

\bibitem{qi2018geonet}
Xiaojuan Qi, Renjie Liao, Zhengzhe Liu, Raquel Urtasun, and Jiaya Jia.
\newblock Geonet: Geometric neural network for joint depth and surface normal
  estimation.
\newblock In {\em CVPR}, pages 283--291, 2018.

\bibitem{romanoni2019tapa}
Andrea Romanoni and Matteo Matteucci.
\newblock Tapa-mvs: Textureless-aware patchmatch multi-view stereo.
\newblock In {\em ICCV}, pages 10413--10422, 2019.

\bibitem{schonberger2016structure}
Johannes~L Schonberger and Jan-Michael Frahm.
\newblock Structure-from-motion revisited.
\newblock In {\em CVPR}, pages 4104--4113, 2016.

\bibitem{schonberger2016pixelwise}
Johannes~L Sch{\"o}nberger, Enliang Zheng, Jan-Michael Frahm, and Marc
  Pollefeys.
\newblock Pixelwise view selection for unstructured multi-view stereo.
\newblock In {\em ECCV}, pages 501--518. Springer, 2016.

\bibitem{seki2016patch}
Akihito Seki and Marc Pollefeys.
\newblock Patch based confidence prediction for dense disparity map.
\newblock In {\em BMVC}, volume~2, page~4, 2016.

\bibitem{sinha2020deltas}
Ayan Sinha, Zak Murez, James Bartolozzi, Vijay Badrinarayanan, and Andrew
  Rabinovich.
\newblock Deltas: Depth estimation by learning triangulation and densification
  of sparse points.
\newblock In {\em ECCV}, 2020.

\bibitem{sun2003stereo}
Jian Sun, Nan-Ning Zheng, and Heung-Yeung Shum.
\newblock Stereo matching using belief propagation.
\newblock {\em PAMI}, 25(7):787--800, 2003.

\bibitem{valentin2018depth}
Julien Valentin, Adarsh Kowdle, Jonathan~T Barron, Neal Wadhwa, Max Dzitsiuk,
  Michael Schoenberg, Vivek Verma, Ambrus Csaszar, Eric Turner, Ivan
  Dryanovski, et~al.
\newblock Depth from motion for smartphone ar.
\newblock {\em ACM Transactions on Graphics (ToG)}, 37(6):1--19, 2018.

\bibitem{vogiatzis2005multi}
George Vogiatzis, Philip~HS Torr, and Roberto Cipolla.
\newblock Multi-view stereo via volumetric graph-cuts.
\newblock In {\em CVPR}, volume~2, pages 391--398, 2005.

\bibitem{wang2018mvdepthnet}
Kaixuan Wang and Shaojie Shen.
\newblock Mvdepthnet: Real-time multiview depth estimation neural network.
\newblock In {\em 3DV}, pages 248--257. IEEE, 2018.

\bibitem{xu2019multi}
Qingshan Xu and Wenbing Tao.
\newblock Multi-scale geometric consistency guided multi-view stereo.
\newblock In {\em CVPR}, pages 5483--5492, 2019.

\bibitem{xu2020planar}
Qingshan Xu and Wenbing Tao.
\newblock Planar prior assisted patchmatch multi-view stereo.
\newblock In {\em AAAI}, volume~34, pages 12516--12523, 2020.

\bibitem{yamaguchi2012continuous}
Koichiro Yamaguchi, Tamir Hazan, David McAllester, and Raquel Urtasun.
\newblock Continuous markov random fields for robust stereo estimation.
\newblock In {\em ECCV}, pages 45--58. Springer, 2012.

\bibitem{yao2018mvsnet}
Yao Yao, Zixin Luo, Shiwei Li, Tian Fang, and Long Quan.
\newblock Mvsnet: Depth inference for unstructured multi-view stereo.
\newblock In {\em ECCV}, pages 767--783, 2018.

\bibitem{yin2019enforcing}
Wei Yin, Yifan Liu, Chunhua Shen, and Youliang Yan.
\newblock Enforcing geometric constraints of virtual normal for depth
  prediction.
\newblock In {\em ICCV}, pages 5684--5693, 2019.

\bibitem{yoon2006adaptive}
Kuk-Jin Yoon and In~So Kweon.
\newblock Adaptive support-weight approach for correspondence search.
\newblock {\em PAMI}, 28(4):650--656, 2006.

\bibitem{yu2017dilated}
Fisher Yu, Vladlen Koltun, and Thomas Funkhouser.
\newblock Dilated residual networks.
\newblock In {\em CVPR}, pages 472--480, 2017.

\bibitem{zhang2018deep}
Yinda Zhang and Thomas Funkhouser.
\newblock Deep depth completion of a single rgb-d image.
\newblock In {\em CVPR}, pages 175--185, 2018.

\end{thebibliography}
}

\clearpage
\section*{Appendix}
\appendix
In this document, we provide a list of supplementary materials that accompany the main paper. 

\section{Detailed Derivations of the Proposed Solver}
\subsection{Preliminaries}
As discussed in the main paper, we solve the depth map and normal map with two separate suboptimization steps with respect to the total energy. Each step contains a plane-based propagation with slanted planes. Recall that $P(x, d, n)$ denotes the slanted plane at pixel coordinate $x$ generating by spanning a plane from the corresponding 3D points recovered from $d$ and $x$ and its surface normal $n$. In practice we parameterize the normal to be $n=(a, b, -1)$, which enables closed-form computation in the normal update step (N-step). Let $(p, q, z)$ denotes the 3D coordinate of the points recovered from $d$ and $x=(u, v)^T$ at the frame coordinate system: 

\begin{equation}
\begin{bmatrix}p \\ q \\ z \end{bmatrix}
=K^{-1}\begin{bmatrix} x \\ 1 \end{bmatrix}d
=\begin{bmatrix} \Tilde{u} \\ \Tilde{v} \\ 1\end{bmatrix}d,
\end{equation}

where $K$ is the camera intrinsic parameter, $x=(u, v)^T$ is the 2D pixel coordinate and $(\Tilde{u}, \Tilde{v}, 1)^\top = K^{-1}(u, v, 1)^\top$ is the normalized homogeneous coordinate. Then, the plane equation of $P(x_i, d_i, n_i)$ with $n_i=(a_i, b_i, -1)$ and the recovered 3D points $(p_i, q_i, z_i)$ can be written as follows:
\begin{equation}
    a_i(p-p_i)+b_i(q-q_i)-(z-z_i)=0.
    \label{eq::supp_planeeq_orig}
\end{equation}
At plane-based propagation, the propagated depth $d_{i\rightarrow j}$ ($d_{j\rightarrow i}$) is computed by projecting the slanted plane at $i$ ($j$) onto the pixel $j$ ($i$). We give the derivation of $d_{i\rightarrow j}$ here:

\begin{align}
    & a_i(p_j-p_i) + b_i(q_j-q_i) - (z_j-z_i) = 0
    \label{eq::supp_planeeq} \\
\Leftrightarrow 
    & a_i(\Tilde{u}_j d_{i\rightarrow j} - p_i) + b_i(\Tilde{v}_j d_{i\rightarrow j} - q_i) - (d_{i\rightarrow j} - z_i) = 0 \\
\Leftrightarrow
    & d_{i\rightarrow j} = \frac{a_i p_i + b_i q_i - z_i}{a_i\Tilde{u}_j + b_i \Tilde{v}_j - 1}
    =\frac{a_i\Tilde{u}_i + b_i\Tilde{v}_i - 1}{a_i\Tilde{u}_j + b_i \Tilde{v}_j - 1}d_i 
    \label{eq::supp_propagation}
\end{align}
Here $d_{i\rightarrow j}$ is non-linearly dependent upon the depth $d_i$ and the surface normal $n_i=(a_i, b_i, -1)$. As there exist second-order terms $a_id_i$ and $b_id_i$ in the denominator, a quadratic energy over both $d_i$ and $n_i$ is infeasible even when the algebraic formulation is employed. Thus, closed-form solution cannot be acquired when the data term and plane-based structural term $E_{i\rightarrow j}$\footnote{$E_{j \rightarrow i}$ does not include the surface normal $n_i$ in its formulation, and thus can only be used when only depth map is required to be solved.} are jointly optimized over $d_i$ and $n_i$. This motivates us to employ iterative suboptimization in the solver to acquire close-form solution, which can further benefit our deep MVS system with end-to-end joint training, as discussed in the main paper.

Before introducing the details of the two update steps of the proposed solver, let us take a step further on the formulation of jointly solving depths and surface normals. We want to note that it is possible to formulate closed-form solution by substitution of variables when only the surface normal data term is employed. This can be achieved by parameterizing the plane equation in Eq. (\ref{eq::supp_planeeq_orig}) as $a_ip + b_iq -(z-t_i)=0$, where $t_i=z_i-a_ip_i-b_iq_i=(1-a_i\Tilde{u}_i-b_i\Tilde{v}_i)d_i$. When the depth data term is not included, by employing the algebraic form we can get a 3x3 linear system with respect to $a_i$, $b_i$ and $t_i$. However, we empirically observe that the depth data term is extremely beneficial in practice.

\subsection{Closed-form Solution}
As discussed in the main paper, we employ suboptimization over the depth map and the surface normal map iteratively. This enables closed-form solution in both steps.

\paragraph{Depth Update (D-step). } At the depth update step (D-step), we fix the surface normal map and solve for the optimal depth map $d^*$. L2 distance between the optimized depth and the propagated depth $d_{j \rightarrow i}$ from neighboring pixels are used in the plane-based structural term. The objective is written as follows (Eqs. (4)(5) in the main paper):

\begin{align}
    & \min_d E_{total} = \min_d E_{d} \\
    & E_{d} = \alpha \sum_i c_i(d_i - \hat{d}_i)^2 + \sum_i \sum_{j \in N(i)}c_jw_{ij}(d_i - d_{j\rightarrow i})^2. \label{eq::supp_Ed}
\end{align}
As discussed in the main paper, we assume fixed neighborhoods to enable parallelization of the solver. Thus, the propagated depth $d_{j \rightarrow i}$ is the projection of the plane $P(x_j, \hat{d}_j, \hat{n}_j)$ at pixel $i$:
\begin{equation}
    d_{j \rightarrow i} = \frac{\hat{a}_j\Tilde{u}_j + \hat{b}_j\Tilde{v}_j - 1}{\hat{a}_j\Tilde{u}_i + \hat{b}_j \Tilde{v}_i - 1}\hat{d}_j 
\end{equation}
Set the first-order derivative to zero we can easily derive the optimal depth $d_i^*$ for each pixel: 
\begin{equation}
 d_i^* = \frac{\alpha c_i \hat{d}_i + \sum_{j\in N(i)} c_iw_{ij}d_{j \rightarrow i}}{\alpha c_i + \sum_{j\in N(i)} c_iw_{ij}}.
\end{equation}

\paragraph{Surface Normal Update (N-step).} At the surface normal update step, we fix the depth map and solve for the optimal surface normal $n^*$. The objective is written as follows (Eqs. (6)(7) in the main paper):

\begin{align}
    & \min_n E_{total} = \min_n E_{n} \\
    &
    \begin{aligned}
    E_{n} &= \alpha \sum_i c_i||n_i - \hat{n}_i||^2 \\
    &+ \sum_i \sum_{j \in N(i)} c_jw_{ij}D_n(d_j, P(x_i, d_i, n_i)).
    \end{aligned}
    \label{eq::supp_En}
\end{align}
Here $D_n$ is a distance function defined over $d_j$ and the plane $P(x_i, d_i, n_i)$ being optimized. Note that because $d_{i \rightarrow j}$ is non-linearly dependent over $n_i$ as shown in Eq. (\ref{eq::supp_propagation}) in this supplementary material, we cannot directly use L2 distance as in the D-step. Instead, we employ the algebraic formulation of the plane equation and directly formulate the distance function $D_n$ as the square of the LHS of Eq. (\ref{eq::supp_planeeq}) in this supplementary material:

\begin{equation}
    D_n(d_j, P(x_i, d_i, n_i)) = [a_i(p_j-p_i) + b_i(q_j-q_i)-(z_j-z_i))]^2.
\end{equation}
Because the depth map is fixed in the N-step, the only two unknown variables are $a_i$ and $b_i$, which represent the surface normal $n_i=(a_i, b_i, -1)$. Note that this parameterization is feasible because all visible surfaces are facing the position where the camera center locates. However, numerical problems may occur when there exist ill-posed cases with surfaces that are nearly parallel to the corresponding camera rays. Thus, we clip the absolute value of the solved $a_i$ and $b_i$ with a threshold 20.0. This operation is empirically crucial to stabilize the end-to-end training process. 

By setting the first-order derivatives to zero we can get a 2x2 linear system over $a_i^*$ and $b_i^*$, where the optimal surface normal $n_i^*$ is parameterized with $n_i^*=(a_i^*, b_i^*, -1)$. The coefficients are listed as follows:

\begin{align}
    &\begin{bmatrix} A_{11} & A_{12} \\ A_{21} & A_{22}\end{bmatrix} \begin{bmatrix}a_i^* \\ b_i^*\end{bmatrix}=
    \begin{bmatrix}B_1 \\ B_2 \end{bmatrix} \\
    & A_{11} = \alpha c_i + \sum_{j \in N(i)}c_jw_{ij}(p_j-p_i)^2 \\
    & A_{22} = \alpha c_i + \sum_{j \in N(i)}c_jw_{ij}(q_j-q_i)^2 \\
    & A_{12} = A_{21} = \sum_{j \in N(i)}c_jw_{ij}(p_j-p_i)(q_j-q_i) \\
    & B_1 = \alpha c_i \hat{a}_i + \sum_{j \in N(i)}c_jw_{ij}(p_j-p_i)(z_j-z_i) \\
    & B_2 = \alpha c_i \hat{b}_i + \sum_{j \in N(i)}c_jw_{ij}(q_j-q_i)(z_j-z_i)
\end{align}

In the final step, normalization is applied on the output surface normal to get the final prediction with standard parameterization: 
\begin{equation}
    n=(\frac{a}{\sqrt{a^2 + b^2 + 1}}, \frac{b}{\sqrt{a^2 + b^2 + 1}}, -\frac{1}{\sqrt{a^2 + b^2 + 1}}).
\end{equation}

\subsection{Extension: Incorporating Customized Depth Confidence and Normal Confidence}
While we present all the derivations with a single confidence map $c_i$ for simplicity, a separate depth confidence map $c_d$ and surface normal confidence map $c_n$ can be used in practice to further improve the solver. Employing those separate customized confidence maps will change the data term (Eq. (2) in the main paper) into the formulation below:

\begin{equation}
    E_{data} = \sum_{i} c_{d,i}(d_i - \hat{d}_i)^2 + \sum_i c_{n,i}||n_i - \hat{n}_i||^2,
\end{equation}
where $c_{d,i}$ and $c_{n,i}$ denotes the per-pixel depth and surface normal confidence respectively. Furthermore, the confidence used in the plane-based structural term can also be monitored by the separate confidence maps.

\paragraph{Separate Confidence Maps used in the D-step.}
When the depth confidence and normal confidence are used jointly, the final formulation of $E_d$ becomes the form as follows:
\begin{equation}
\begin{aligned}
    E_{d} &= \alpha \sum_i c_{d,i}(d_i - \hat{d}_i)^2 \\
    &+ \sum_i \sum_{j \in N(i)}c_{d,j}c_{n,j}w_{ij}(d_i - d_{j\rightarrow i})^2. 
\end{aligned}
\label{eq::supp_finalEd}
\end{equation}
Note that both the depth confidence $c_{d,j}$ and normal confidence $c_{n,j}$ are used in the plane-based structural term. This is due to the fact that the quality of the propagated depth $d_{j \rightarrow i}$ depends on both $\hat{d}_j$ and $\hat{n}_j$. 
\paragraph{Separate Confidence Maps used in the N-step.}
As for the surface normal update step, because $E_{i \rightarrow j}$ is employed, $\hat{n}_j$ is non-relevant to the computation so the plane-based structural term will not be affected by $c_{n,j}$. However, since the depth map is fixed here and the computation of $d_{i \rightarrow j}$ depends on the quality of $\hat{d}_i$, we multiply the plane-based structural term with the depth confidence of the studied pixel $c_{d, i}$. The final formulation of $E_{n}$ is written as follows:
\begin{equation}
    \begin{aligned}
    E_{n} &= \alpha \sum_i c_{n,i}||n_i - \hat{n}_i||^2 \\
    &+ \sum_i c_{d,i}\sum_{j \in N(i)} c_{d,j}w_{ij}D_n(d_j, P(x_i, d_i, n_i)).
    \end{aligned}
    \label{eq::supp_finalEn}
\end{equation}

Closed-form solution can be easily derived for the modified objectives in Eq. (\ref{eq::supp_finalEd}) and Eq. (\ref{eq::supp_finalEn}) following the previous discussion.

In our implementation, the depth confidence map and normal confidence map are jointly predicted by the cost-volume based neural networks. The supervision is acquired by computing the relative depth error and normal angle error, as described in the main paper. At inference, the hybrid confidence map combining the deep depth confidence and the geometric confidence is used as the depth confidence map, while the normal confidence only employs the deep normal confidence prediction.

\section{Network Architectures}
We follow prior works \cite{im2019dpsnet, kusupati2020normal} to design the initial depth, surface normal, and confidence estimation networks. 

For the depth and normal branches, we use the same network architectures as \cite{kusupati2020normal} without the consistency module. Specifically, the target and reference images are first encoded to get the feature maps. Then we use plane sweeping with 64 hypothesis planes to build the feature cost volume. The 3D CNNs and 2D context CNNs are applied on the cost volume to aggregate and regularize the cost information. The soft argmin is used to regress the final depth values from the final cost volume. The cost volume information is also utilized for the multi-view surface normal estimation. The intermediate cost volume features are concatenated with the world coordinates of every voxel, and then transformed by several 3D CNNs to get 8 cost volume slices. Each slice is processed by 7 shared layers of 2D convolutions of dilated $3\times 3$ kernels. The output of all slices are summed and normalized to get the final surface normal. We recommend the reader to refer to \cite{kusupati2020normal} for more details.

\begin{figure}[tb]
\includegraphics[width=\linewidth]{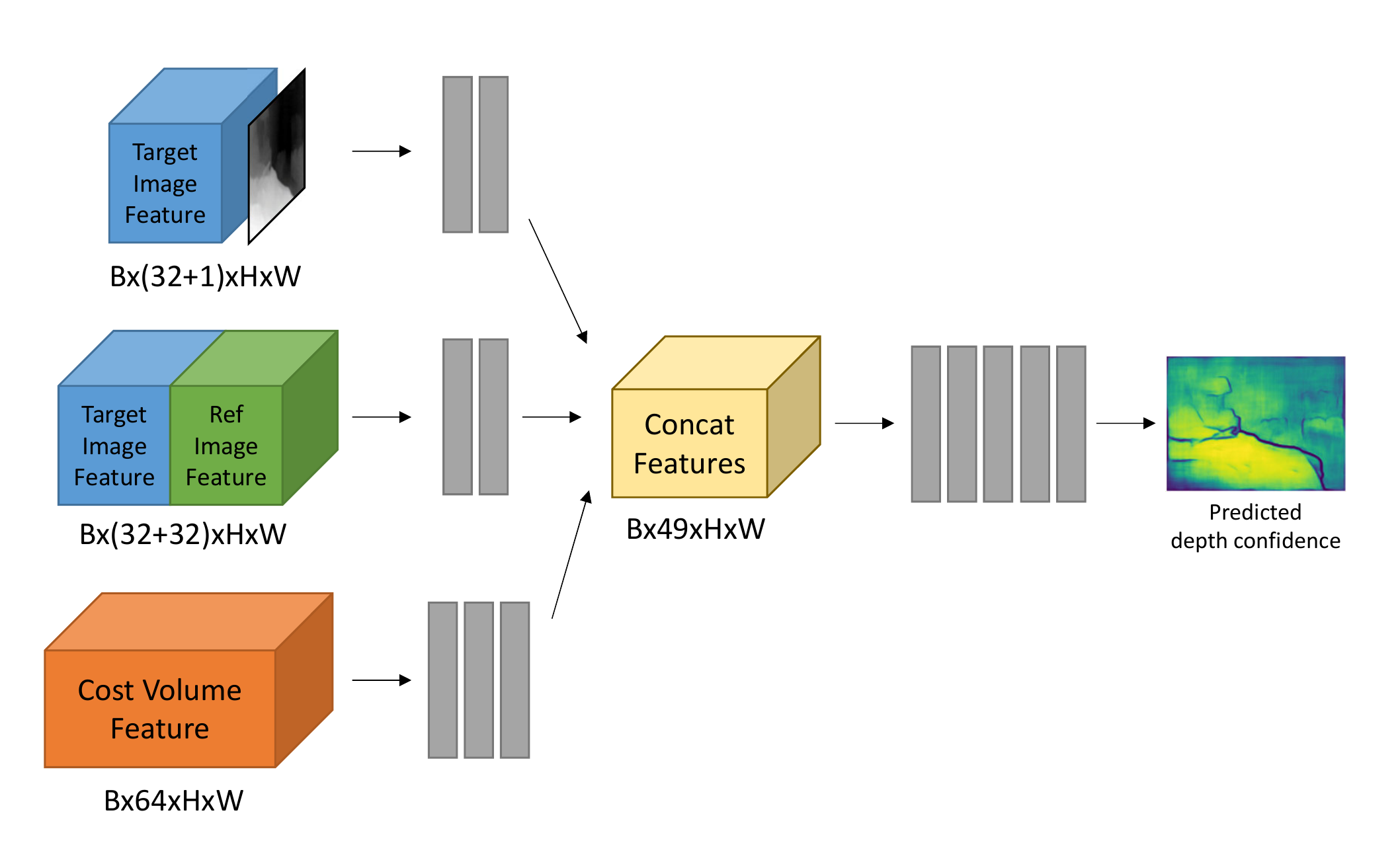}
\centering
\caption{The inputs are processed with three mini-branches (with 2 or 3 layers of 3x3 convolutions), and then jointly fed into 5 dilated convolutions followed by final sigmoid activation to regress the depth confidence map.}
\label{fig::supp_dconf}
\vspace{-10pt}
\end{figure}

\begin{figure}[tb]
\includegraphics[width=\linewidth]{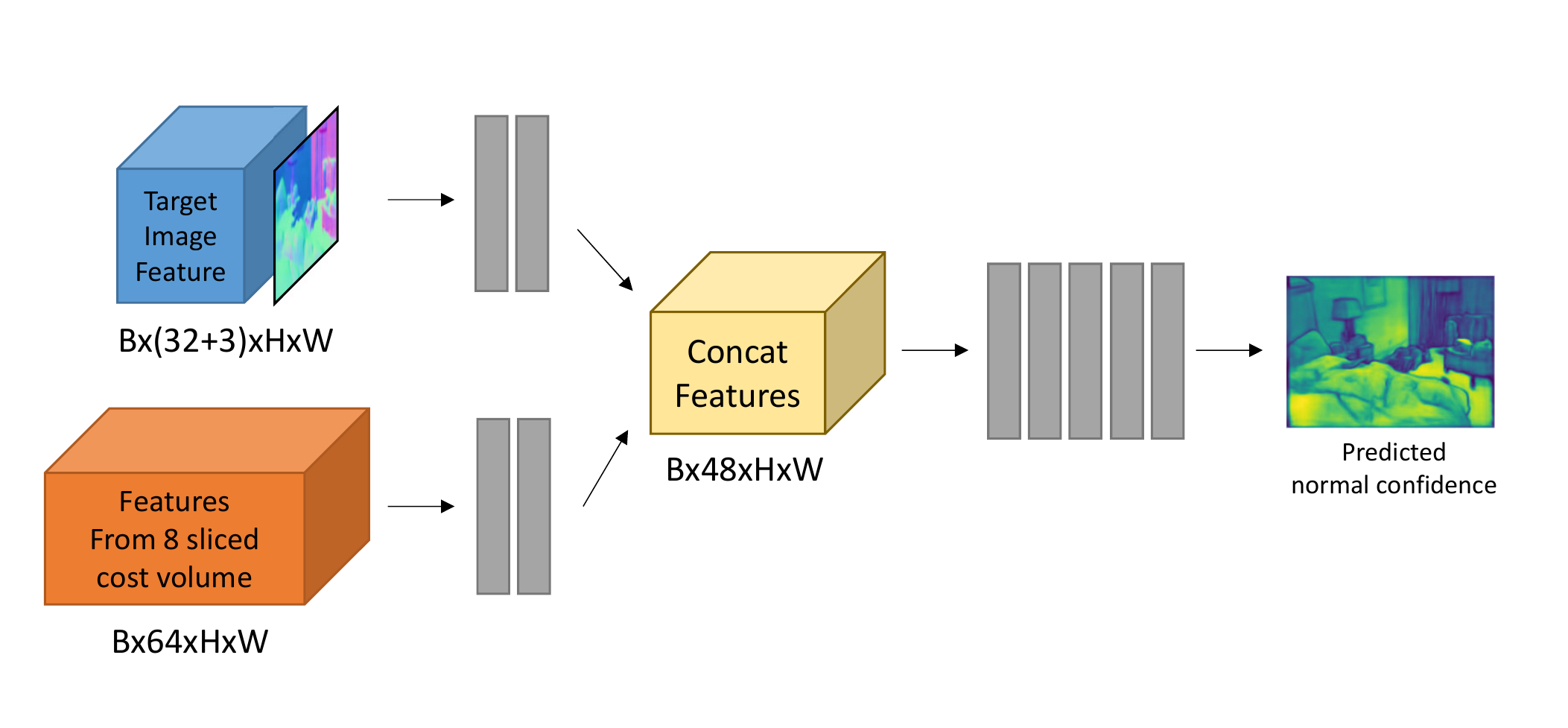}
\centering
\caption{The inputs are processed with two mini-branches (with 2 layers of 3x3 convolutions), and then jointly fed into 5 dilated convolutions followed by final sigmoid activation to regress the normal confidence map.}
\label{fig::supp_nconf}
\vspace{-10pt}
\end{figure}

For the confidence branch, we input multiple sources to the network to better estimate the confidence. To predict the depth confidence, we utilize target image features, homography-warped reference image features using currently predicted depth, cost volume features before softmax, and the predicted depth. As illustrated in Figure \ref{fig::supp_dconf}, these inputs are processed by three mini-branches, and each mini-branch contains two or three layers of $3\times 3$ convolutions. To be specific, the first branch (target image feature + predicted depth) consists of two $3\times 3$ convolution layers whose output channels are both 16. The second branch (target image feature + warped reference image feature) has also two $3\times 3$ convolution layers whose output channels are both 32. The third branch (cost volume feature) consists of three $3\times 3$ convolution layers with [64, 32, 1] as each layer's output channel number. The outputs from three mini-branches are then concatenated, and five dilated $3\times 3$ convolution layers with output channels [64, 64, 64, 32, 1] and dilations [1, 2, 4, 1, 1], followed by the final sigmoid activation, are applied to jointly predict the depth confidence. For surface normal confidence, we use target image features, intermediate features from the previously discussed 8 cost volume slices, and the predicted surface normal, as shown in Figure \ref{fig::supp_nconf}. These inputs are processed by two mini-branches, then concatenated and used to jointly predict the surface normal confidence. The first branch consists of two $3\times 3$ convolution layers with both 16 output channels, and the second branch has also two $3\times 3$ convolution layers with both 32 output channels. Then the outputs from these two mini-branches are concatenated and processed by five dilated $3\times 3$ convolution layers with output channels [64, 64, 64, 32, 1] and dilations [1, 2, 4, 1, 1], plus final sigmoid activation, to get the confidence map for surface normal prediction. 


\section{System Training}
Our proposed deep MVS system is trained with two stages. In the first stage, we train the initial depth, normal, and confidence estimation network for 15 epochs. The training loss for the initial depth, normal and confidence estimation, denoted as $L_{net}$, includes three parts: $L_{net} = w_d L_{d} + w_n L_{n} + w_c L_{c}$, where $w_d, w_n, w_c$ are three hyper-parameters to balance different loss terms. $L_{d}$ and $L_{n}$ are formulated as smoothed L1 loss, and $L_{c}$ employs cross-entropy loss. $L_{c}$ consists of depth confidence loss $L_{cd}$ and surface normal confidence loss $L_{cn}$, and is computed as $L_{c} = L_{cd} + L_{cn}$.

In the second stage, we finetune the network jointly with the proposed iterative depth normal solver for 10 epochs. During the training, We iteratively refine the depth map and surface normal map for 5 times to balance the computation and effectiveness. We apply the depth and surface normal loss on both the initial predictions and the final solved geometry. Thus the total loss for end-to-end training is $L_{total} = \lambda L_{net} + L_{solver}$, where $L_{solver} = w_{d'} L_{d'} + w_{n'} L_{n'}$ is the depth and normal loss for the solved geometry. $L_{d'}$ and $L_{n'}$ are smoothed L1 loss for the refined depth and normal, respectively. $w_{d'}$ and $w_{n'}$ are hyper-parameters. $\lambda$ is the hyper-parameter for weighting these two losses $L_{net}$ and $L_{solver}$. We choose to include the $L_{net}$ in the end-to-end training to regularize the initial depth and normal predictions, which in practice stabilizes the joint training.

\section{Implementation Details}
The hyperparameters are heuristically selected without much tuning. We set loss weights [$\lambda$, $w_d$, $w_n$, $w_c$, $w_{d'}$, $w_{n'}$] to [0.7, 1.0, 3.0, 0.2, 1.0, 3.0]. The scaling factors $\gamma_1$, $\gamma_2$ for depth and normal confidence groundtruth are both set to 5.0 in the training. The spatial and color weights for bilateral affinity $\sigma_x^2, \sigma_c^2$ are set to 2.5 and 25.0, and the weight $\alpha$ for depth and normal data term of the energy is set to 1.0. These hyper-parameters are fixed during both training and inference.

\section{Notations}
\begin{table*}[tb]
\begin{center}

\scalebox{0.8}{
\begin{tabular}{l|l|l}
\hline
Notations & Descriptions & Appearances \\
\hline
$E_{total}$ & The total energy potential for the solver & Sec. 3.1, Sec. 3.2, Sec. A.2 \\
\hline
$E_{data}$ & The data term in the total energy potential & Sec. 3.1, Sec. A.3 \\
\hline
$E_{plane}$ & The plane-based structural term in the total energy potential & Sec. 3.1 \\
\hline
$\alpha$ & The hyperparameter to balance the data term and structural term & Sec. 3.1, Sec. 3.2, Sec. A.2, Sec. A.3, Sec. E \\
\hline
$x, d, n, c$ & 2D coordinate, depth, surface normal, and confidence & Sec. 3.1, Sec. 3.2, Sec. A.1, Sec. A.2, Sec. A.3 \\
\hline
$x_i, d_i, n_i, c_i$ & Per-pixel 2D coordinate, depth, surface normal, and confidence & Sec. 3.1, Sec. 3.2, Sec. A.1, Sec. A.2 \\
\hline
$\hat{d}_i, \hat{n}_i$ & Initial per-pixel depth and surface normal & Sec. 3.1, Sec. 3.2, Sec. A.2, Sec. A.3 \\
\hline
$P(x,d,n)$ & The plane generated by current 2D coordinate $x$, depth $d$, and normal $n$ & Sec. 3.1, Sec. 3.2, Sec. A.1, Sec. A.2, Sec. A.3\\
\hline
$E_{j\rightarrow i}, E_{i\rightarrow j}$ & The plane-based structural term defined in two directions & Sec. 3.1, Sec. 3.2, Sec. A.1, Sec. A.3 \\
\hline
$d_{j\rightarrow i}$ & The projection of the plane $P(x_j, d_j, n_j)$ at pixel $i$  & Sec. 3.1, Sec. 3.2, Sec. A.1, Sec. A.2, Sec. A.3 \\
\hline 
$d_{i\rightarrow j}$ & The projection of the plane $P(x_i, d_i, n_i)$ at pixel $j$  & Sec. 3.1, Sec. 3.2, Sec. A.1, Sec. A.2, Sec. A.3 \\
\hline
$w_{ij}$ & Edge-aware bilateral affinity between pixel $i$ and $j$ & Sec. 3.1, Sec. 3.2, Sec. A.2, Sec. A.3 \\
\hline
$I_i$ & RGB value at pixel $i$ & Sec. 3.1 \\
\hline
$\sigma_x, \sigma_c$ & The hyperparameters for the spatial term and color term in the bilateral affinity & Sec. 3.1, Sec. E \\
\hline
$E_d$ & The minimized objective in the D-step & Sec. 3.2, Sec. A.2, Sec. A.3\\
\hline
$N(i)$ & The defined neighborhoods of pixel $i$ & Sec. 3.2, Sec. A.2, Sec. A.3 \\
\hline
$E_n$ & The minimized objective in the N-step & Sec. 3.2, Sec. A.2, Sec. A.3 \\
\hline
$D_n$ & The distance function between the depth $d$ and slanted plane $P$ used in the N-step & Sec. 3.2, Sec. A.2, Sec. A.3 \\
\hline
$a, b$ & Components of the parameterized surface normal, $n=(a,b,-1)$ & Sec. 3.2, Sec. A.1, Sec. A.2 \\
\hline
$c_{dgt}, c_{ngt}$ & Groundtruth (GT) confidence maps for depth and surface normal & Sec. 4.1 \\
\hline
$e_{rel}, e_{ang}$ & Relative depth error and normal angle error between predictions and groundtruths & Sec. 4.1 \\
\hline
$\gamma_1, \gamma_2$ & Hyperparameters used in the computation of the GT depth and normal confidence & Sec. 4.1, Sec. E \\
\hline
$p, q, z$ & The 3D coordinate of the unprojected point & Sec. A.1, Sec. A.2 \\
\hline
$K$ & Camera intrinsic matrix & Sec. A.1 \\
\hline
$u, v$ & 2D pixel coordinates, $x = (u, v)^T$ & Sec. A.1 \\
\hline
$\Tilde{u}, \Tilde{v}$ & $(\Tilde{u}, \Tilde{v}, 1)^T=K^{-1}(u, v, 1)^T$ & Sec. A.1 \\
\hline
$d^*$ & The optimal depth map in the D-step & Sec. A.2 \\
\hline
$n^*$ & The optimal surface normal map in the N-step & Sec. A.2 \\
\hline
$a^*, b^*$ & Components of the optimal surface normal $n^*$, $n^* = (a^*, b^*, -1)$ & Sec. A.2 \\
\hline
$A_{11}, A_{12}, A_{21}, A_{22},$ & {Coefficients used in the N-step computation} & {Sec. A.2} \\
$B_1, B_2$ & & \\
\hline
$c_d$, $c_n$ & The depth confidence map and surface normal confidence map & Sec. A.3 \\
\hline
$c_{d,i}$, $c_{n,i}$ & Per-pixel depth confidence and surface normal confidence & Sec. A.3 \\
\hline
$L_{net}$ & Training loss for the initial depth, surface normal, and confidence network & Sec. D \\
\hline
$L_d, L_n, L_c$ & Losses of initial depth, surface normal and confidence & Sec. D \\
\hline
$w_d, w_n, w_c$ & Loss weights of initial depth, surface normal and confidence & Sec. D, Sec. E \\
\hline
$L_{cd}, L_{cn}$ & Confidence loss for depth and normal, $L_c = L_{cd} + L_{cn}$ & Sec. D \\
\hline
$L_{total}$ & Total training loss in the end-to-end training & Sec. D \\
\hline
$L_{solver}$ & The Loss defined over the solved geometry $L_{solver} = w_{d'} L_{d'} + w_{n'} L_{n'}$ & Sec. D \\
\hline
$L_{d'}, L_{n'}$ & Losses of the solved depth and surface normal & Sec. D \\
\hline
$w_{d'}, w_{n'}$ & Loss weights of the solved depth and surface normal & Sec. D, Sec. E \\
\hline
$\lambda$ & The hyperparameter used to balance $L_{net}$ and $L_{solver}$ & Sec. D, Sec. E \\
\hline

\end{tabular}}

\vspace{0.1cm}
\caption{Notations used in the main paper and supplementary material.}
\label{tab::supp_notation}
\end{center}
\vspace{-20pt}
\end{table*}
We provide a notation cheat sheet in Table \ref{tab::supp_notation}, which describes the relevant notations used in the main paper and this supplementary material.

\section{Additional Visualization}
\begin{figure*}[tb]
\scriptsize
\setlength\tabcolsep{1.0pt} 
\begin{tabular}{cccccccc}
{\includegraphics[width=0.115\linewidth]{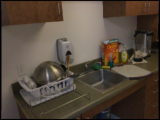}} & 
{\includegraphics[width=0.115\linewidth]{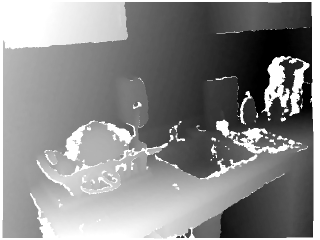}} &
{\includegraphics[width=0.115\linewidth]{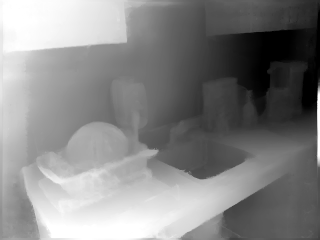}} & 
{\includegraphics[width=0.115\linewidth]{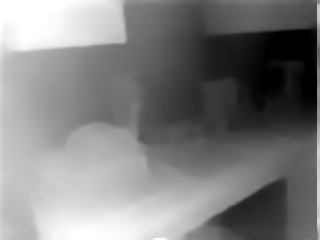}} & 
{\includegraphics[width=0.115\linewidth]{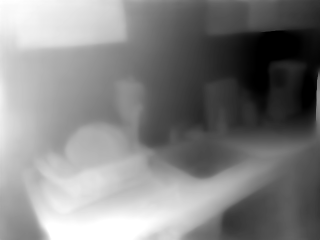}} &
{\includegraphics[width=0.115\linewidth]{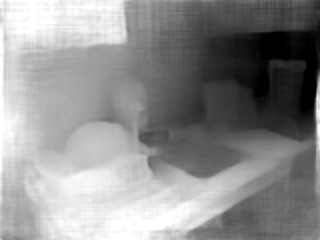}} &
{\includegraphics[width=0.115\linewidth]{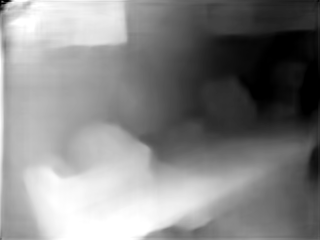}} &
{\includegraphics[width=0.115\linewidth]{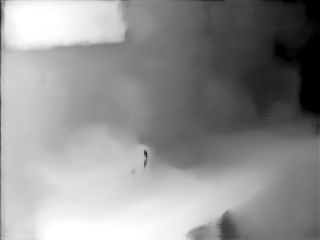}} \\

{\includegraphics[width=0.115\linewidth]{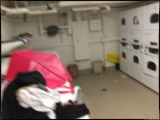}} & 
{\includegraphics[width=0.115\linewidth]{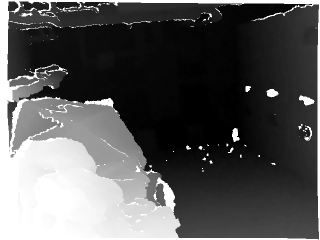}} &
{\includegraphics[width=0.115\linewidth]{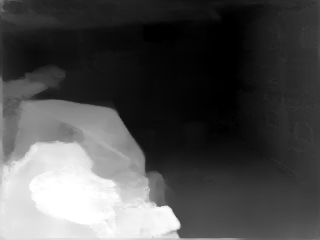}} & 
{\includegraphics[width=0.115\linewidth]{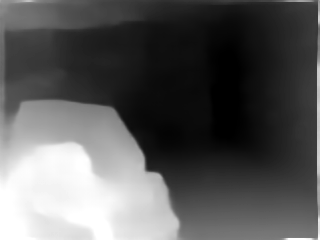}} & 
{\includegraphics[width=0.115\linewidth]{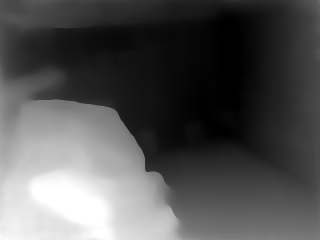}} &
{\includegraphics[width=0.115\linewidth]{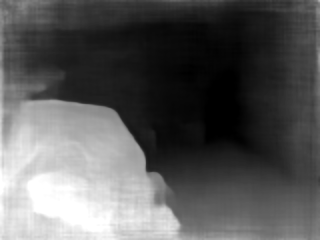}} &
{\includegraphics[width=0.115\linewidth]{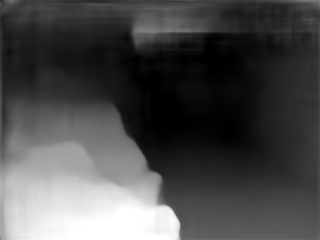}} &
{\includegraphics[width=0.115\linewidth]{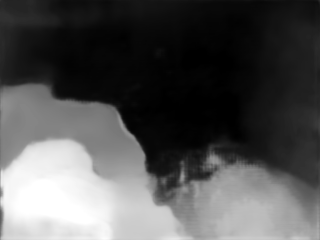}} \\

{\includegraphics[width=0.115\linewidth]{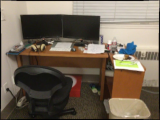}} & 
{\includegraphics[width=0.115\linewidth]{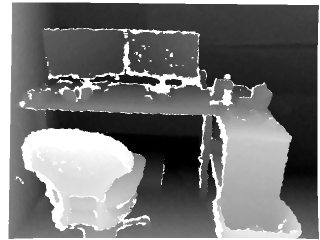}} &
{\includegraphics[width=0.115\linewidth]{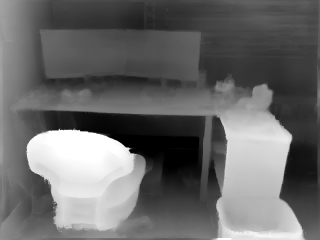}} & 
{\includegraphics[width=0.115\linewidth]{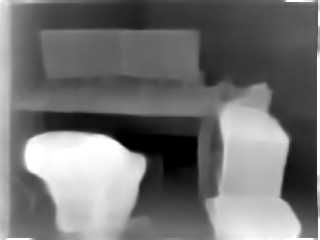}} & 
{\includegraphics[width=0.115\linewidth]{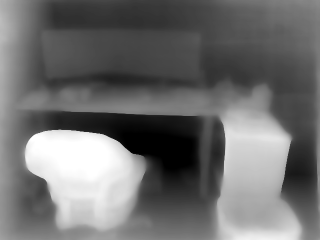}} &
{\includegraphics[width=0.115\linewidth]{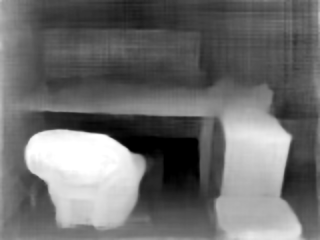}} &
{\includegraphics[width=0.115\linewidth]{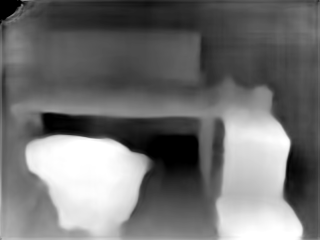}} &
{\includegraphics[width=0.115\linewidth]{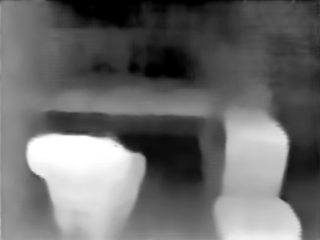}} \\

{\includegraphics[width=0.115\linewidth]{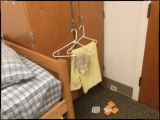}} & 
{\includegraphics[width=0.115\linewidth]{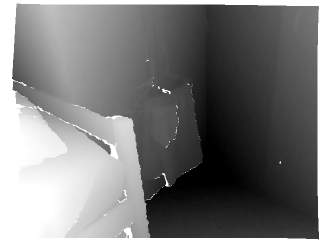}} &
{\includegraphics[width=0.115\linewidth]{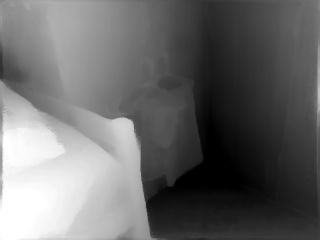}} & 
{\includegraphics[width=0.115\linewidth]{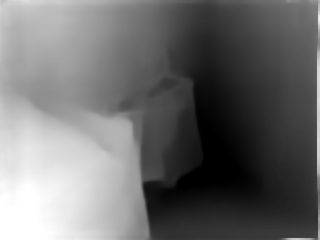}} & 
{\includegraphics[width=0.115\linewidth]{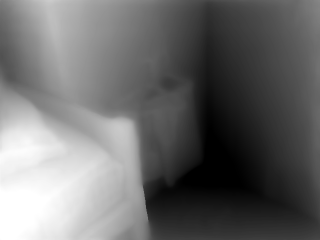}} &
{\includegraphics[width=0.115\linewidth]{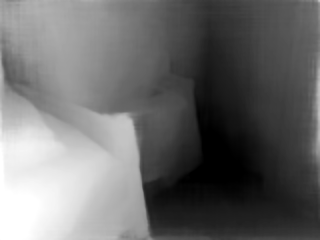}} &
{\includegraphics[width=0.115\linewidth]{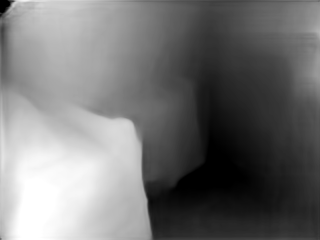}} &
{\includegraphics[width=0.115\linewidth]{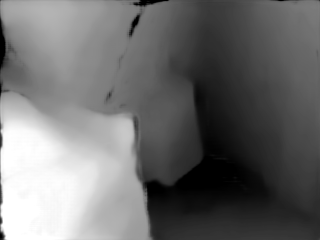}} \\

{\includegraphics[width=0.115\linewidth]{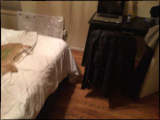}} & 
{\includegraphics[width=0.115\linewidth]{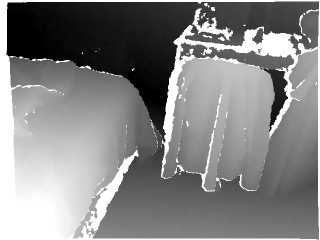}} &
{\includegraphics[width=0.115\linewidth]{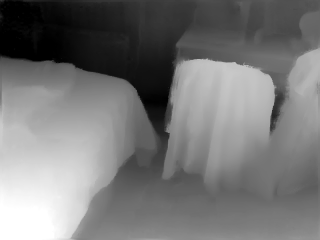}} & 
{\includegraphics[width=0.115\linewidth]{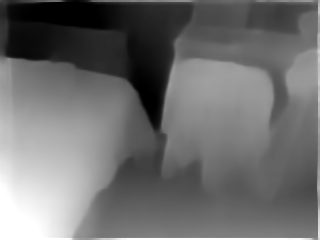}} & 
{\includegraphics[width=0.115\linewidth]{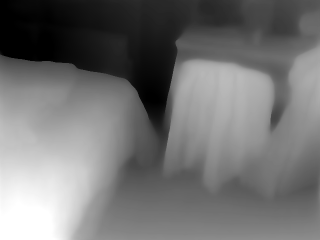}} &
{\includegraphics[width=0.115\linewidth]{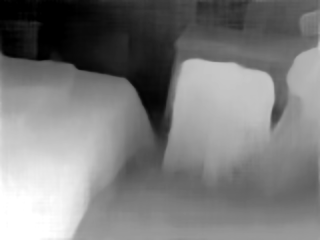}} &
{\includegraphics[width=0.115\linewidth]{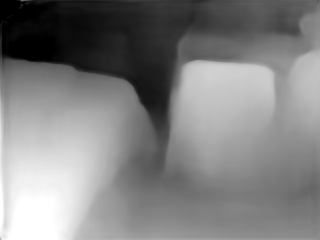}} &
{\includegraphics[width=0.115\linewidth]{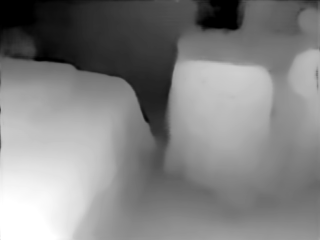}} \\

{\includegraphics[width=0.115\linewidth]{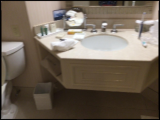}} & 
{\includegraphics[width=0.115\linewidth]{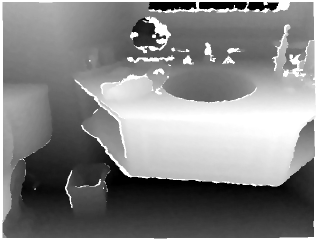}} &
{\includegraphics[width=0.115\linewidth]{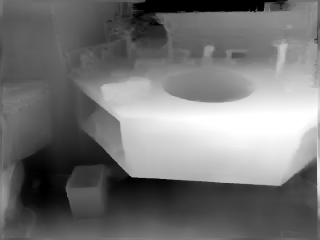}} & 
{\includegraphics[width=0.115\linewidth]{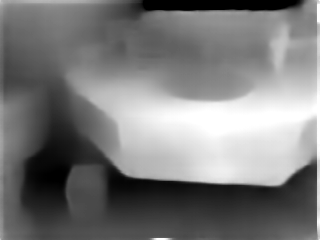}} & 
{\includegraphics[width=0.115\linewidth]{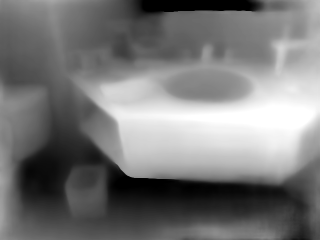}} &
{\includegraphics[width=0.115\linewidth]{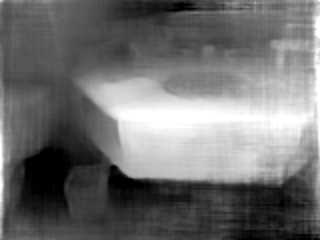}} &
{\includegraphics[width=0.115\linewidth]{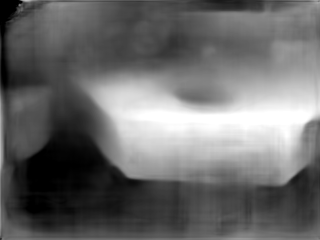}} &
{\includegraphics[width=0.115\linewidth]{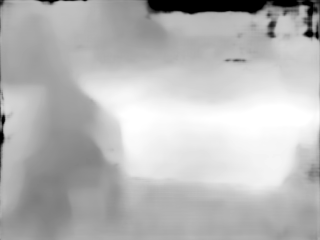}} \\

{\includegraphics[width=0.115\linewidth]{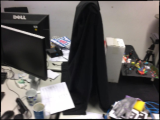}} & 
{\includegraphics[width=0.115\linewidth]{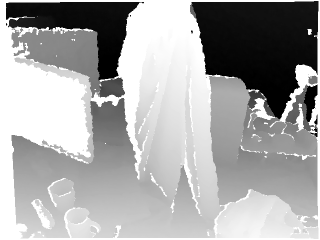}} &
{\includegraphics[width=0.115\linewidth]{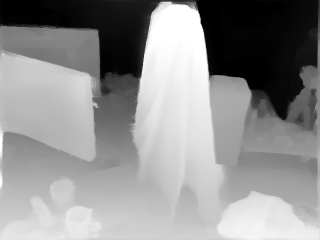}} & 
{\includegraphics[width=0.115\linewidth]{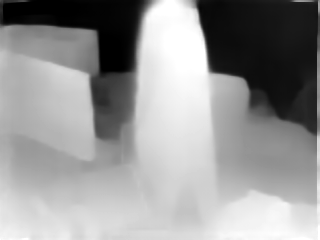}} & 
{\includegraphics[width=0.115\linewidth]{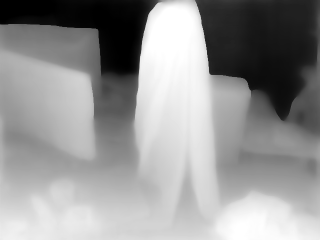}} &
{\includegraphics[width=0.115\linewidth]{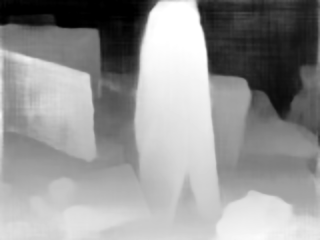}} &
{\includegraphics[width=0.115\linewidth]{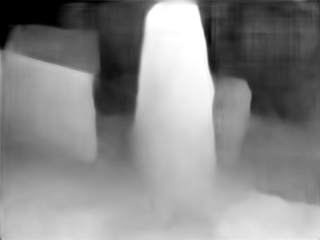}} &
{\includegraphics[width=0.115\linewidth]{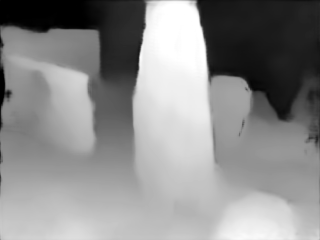}} \\

{\includegraphics[width=0.115\linewidth]{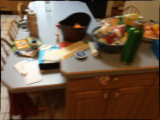}} & 
{\includegraphics[width=0.115\linewidth]{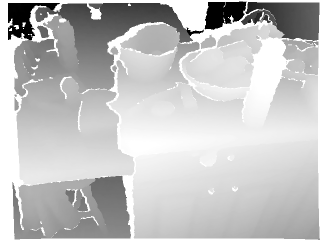}} &
{\includegraphics[width=0.115\linewidth]{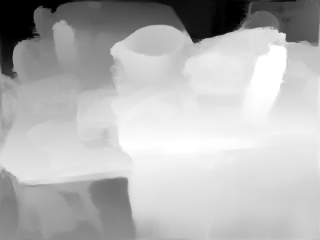}} & 
{\includegraphics[width=0.115\linewidth]{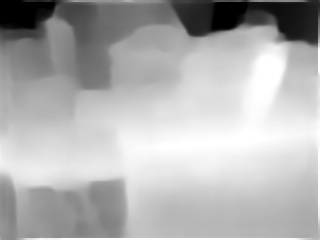}} & 
{\includegraphics[width=0.115\linewidth]{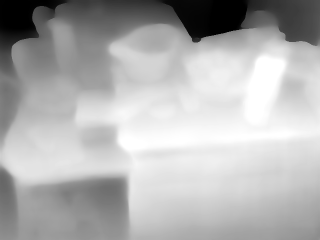}} &
{\includegraphics[width=0.115\linewidth]{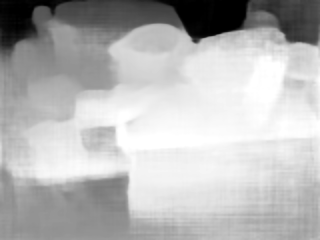}} &
{\includegraphics[width=0.115\linewidth]{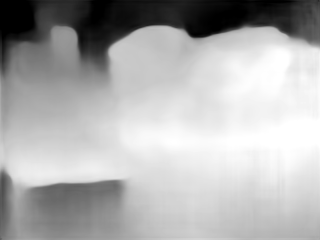}} &
{\includegraphics[width=0.115\linewidth]{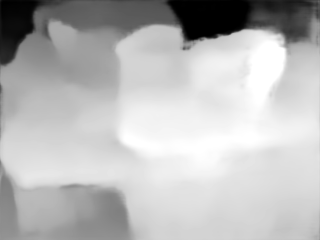}} \\

{\includegraphics[width=0.115\linewidth]{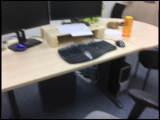}} & 
{\includegraphics[width=0.115\linewidth]{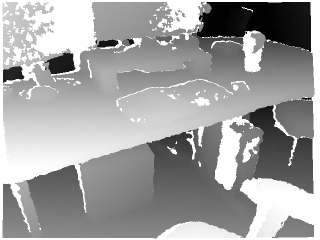}} &
{\includegraphics[width=0.115\linewidth]{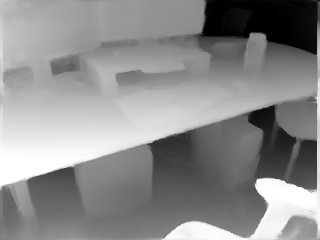}} & 
{\includegraphics[width=0.115\linewidth]{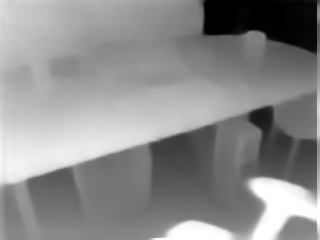}} & 
{\includegraphics[width=0.115\linewidth]{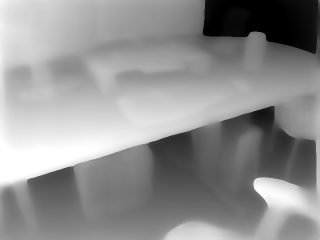}} &
{\includegraphics[width=0.115\linewidth]{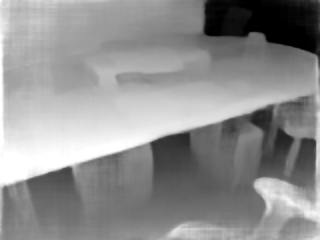}} &
{\includegraphics[width=0.115\linewidth]{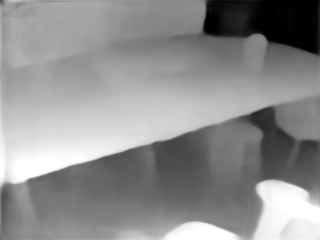}} &
{\includegraphics[width=0.115\linewidth]{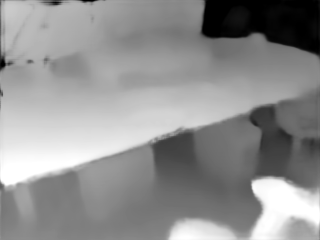}} \\

{\includegraphics[width=0.115\linewidth]{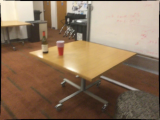}} & 
{\includegraphics[width=0.115\linewidth]{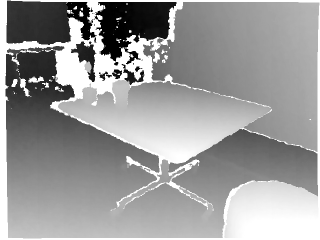}} &
{\includegraphics[width=0.115\linewidth]{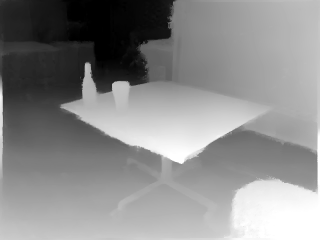}} & 
{\includegraphics[width=0.115\linewidth]{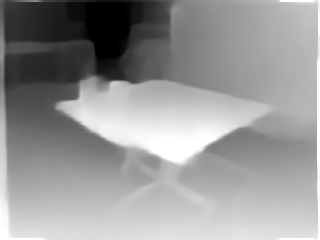}} & 
{\includegraphics[width=0.115\linewidth]{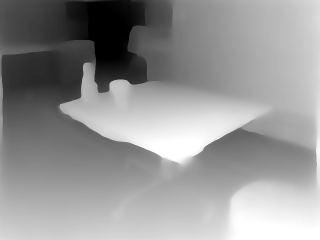}} &
{\includegraphics[width=0.115\linewidth]{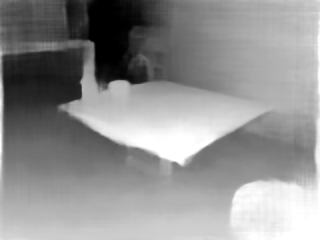}} &
{\includegraphics[width=0.115\linewidth]{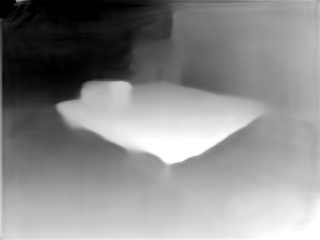}} &
{\includegraphics[width=0.115\linewidth]{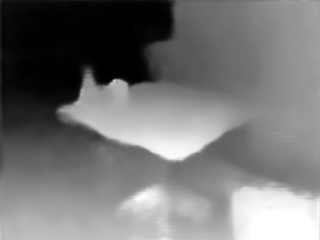}} \\

{\includegraphics[width=0.115\linewidth]{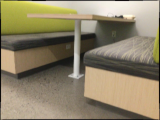}} & 
{\includegraphics[width=0.115\linewidth]{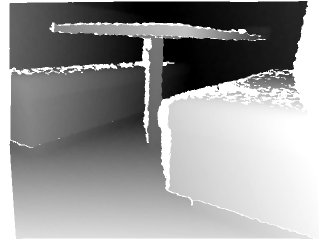}} &
{\includegraphics[width=0.115\linewidth]{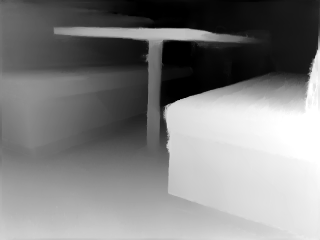}} & 
{\includegraphics[width=0.115\linewidth]{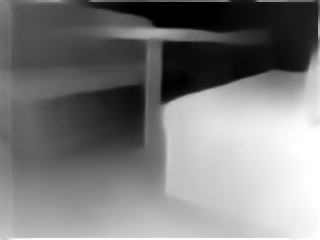}} & 
{\includegraphics[width=0.115\linewidth]{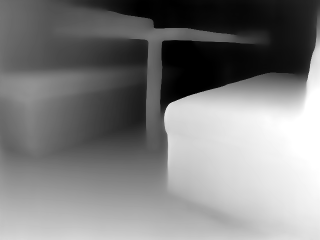}} &
{\includegraphics[width=0.115\linewidth]{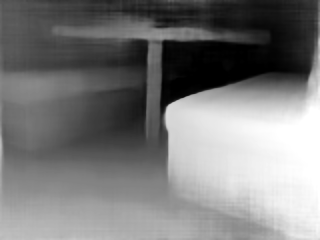}} &
{\includegraphics[width=0.115\linewidth]{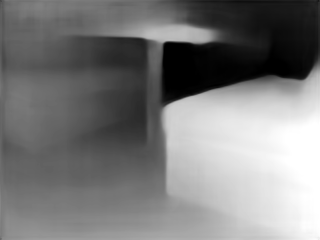}} &
{\includegraphics[width=0.115\linewidth]{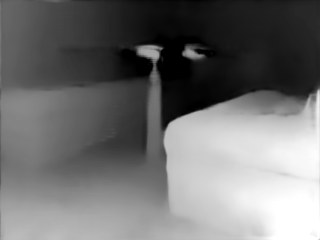}} \\

{\includegraphics[width=0.115\linewidth]{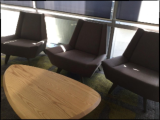}} & 
{\includegraphics[width=0.115\linewidth]{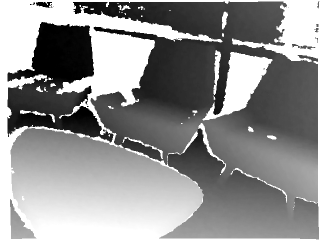}} &
{\includegraphics[width=0.115\linewidth]{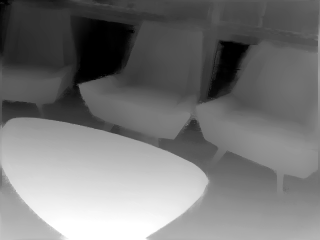}} & 
{\includegraphics[width=0.115\linewidth]{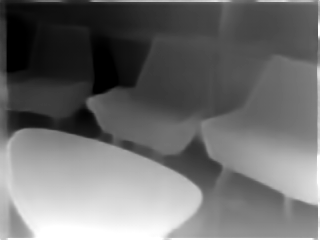}} & 
{\includegraphics[width=0.115\linewidth]{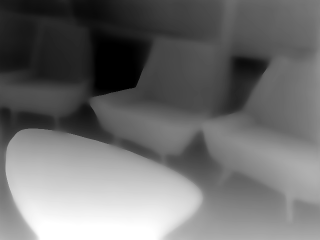}} &
{\includegraphics[width=0.115\linewidth]{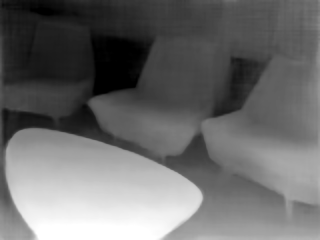}} &
{\includegraphics[width=0.115\linewidth]{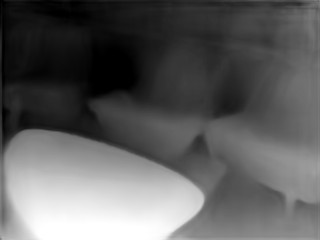}} &
{\includegraphics[width=0.115\linewidth]{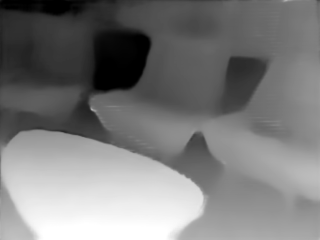}} \\

(a) Image & (b) Groundtruth & (c) Ours & (d) DELTAS \cite{sinha2020deltas} & (e) NAS \cite{kusupati2020normal} & (f) DPSNet (FT) \cite{im2019dpsnet} &  (g) MVDepth (FT) \cite{wang2018mvdepthnet} & (h) N-RGBD \cite{liu2019neural} \\

\end{tabular}
\vspace{0.05cm}
\centering
\caption{More qualititative results of depth estimation on ScanNet \cite{dai2017scannet}. \textbf{Better viewed when zoomed in.}}
\label{fig::supp_depth}
\vspace{-10pt}
\end{figure*}

\begin{figure*}[tb]
\scriptsize
\setlength\tabcolsep{2.0pt} 
\begin{tabular}{ccccc}
{\includegraphics[width=0.18\linewidth]{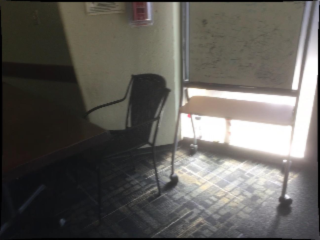}} & 
{\includegraphics[width=0.18\linewidth]{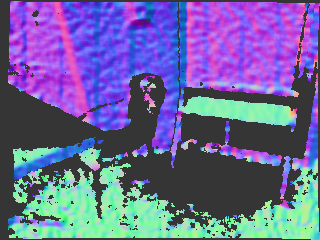}} &
{\includegraphics[width=0.18\linewidth]{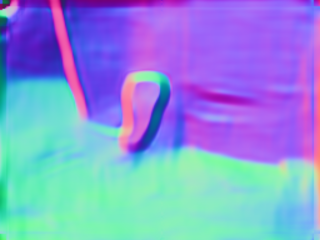}} &
{\includegraphics[width=0.18\linewidth]{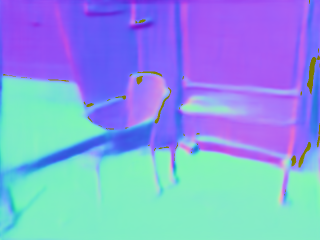}} &
{\includegraphics[width=0.18\linewidth]{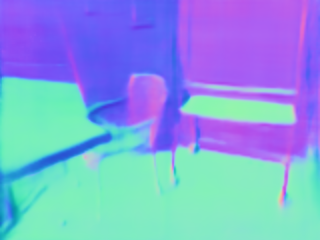}}
\\
{\includegraphics[width=0.18\linewidth]{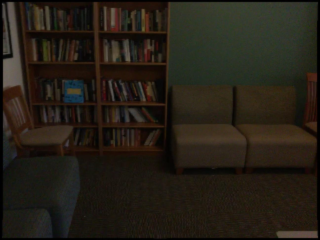}} & 
{\includegraphics[width=0.18\linewidth]{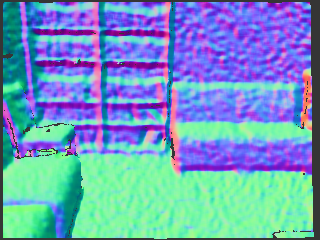}} &
{\includegraphics[width=0.18\linewidth]{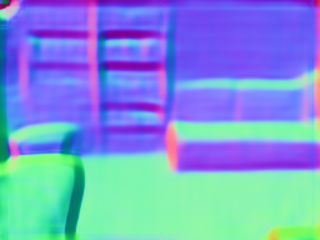}} &
{\includegraphics[width=0.18\linewidth]{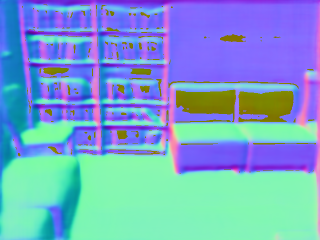}} &
{\includegraphics[width=0.18\linewidth]{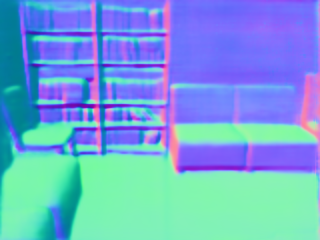}}
\\
{\includegraphics[width=0.18\linewidth]{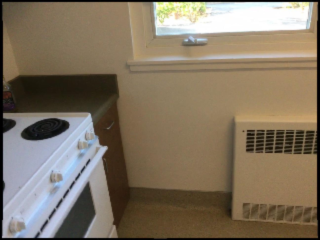}} & 
{\includegraphics[width=0.18\linewidth]{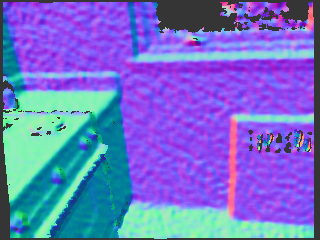}} &
{\includegraphics[width=0.18\linewidth]{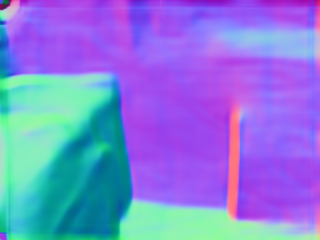}} &
{\includegraphics[width=0.18\linewidth]{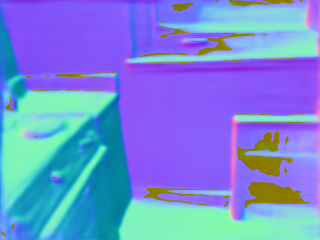}} &
{\includegraphics[width=0.18\linewidth]{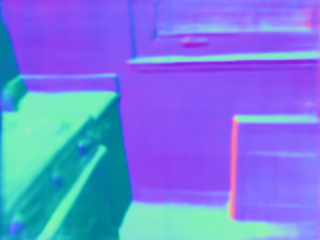}}
\\
{\includegraphics[width=0.18\linewidth]{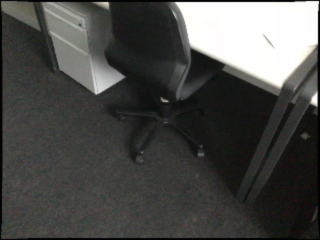}} & 
{\includegraphics[width=0.18\linewidth]{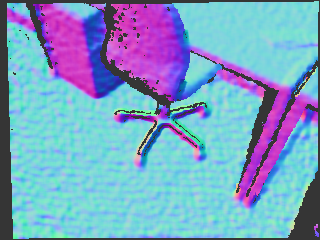}} &
{\includegraphics[width=0.18\linewidth]{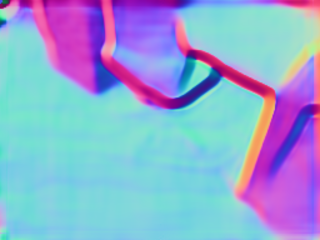}} &
{\includegraphics[width=0.18\linewidth]{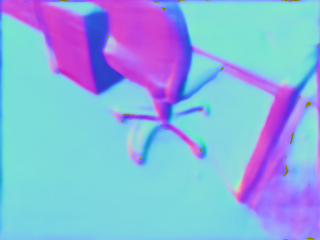}} &
{\includegraphics[width=0.18\linewidth]{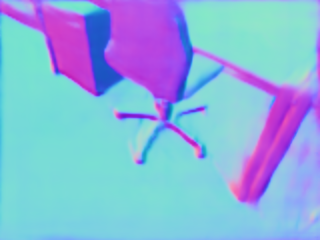}}
\\
{\includegraphics[width=0.18\linewidth]{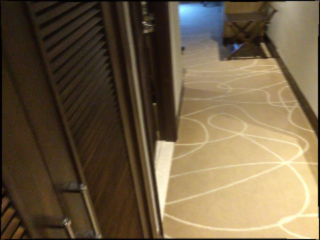}} & 
{\includegraphics[width=0.18\linewidth]{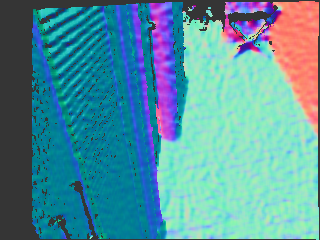}} &
{\includegraphics[width=0.18\linewidth]{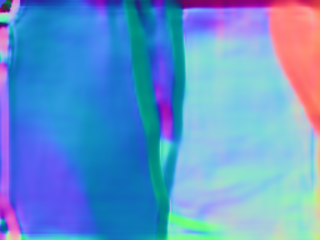}} &
{\includegraphics[width=0.18\linewidth]{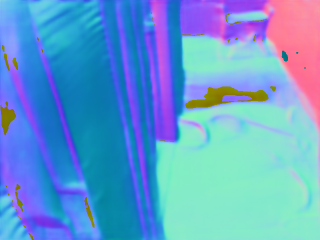}} &
{\includegraphics[width=0.18\linewidth]{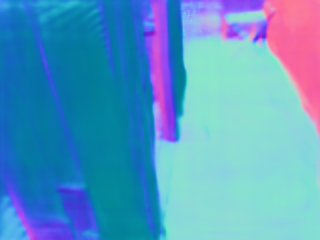}}
\\
{\includegraphics[width=0.18\linewidth]{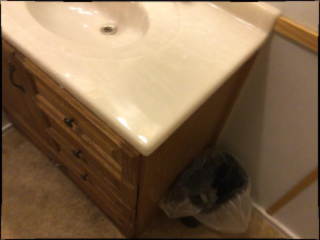}} & 
{\includegraphics[width=0.18\linewidth]{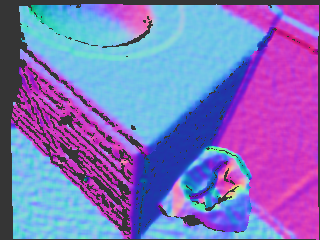}} &
{\includegraphics[width=0.18\linewidth]{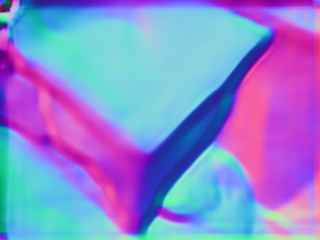}} &
{\includegraphics[width=0.18\linewidth]{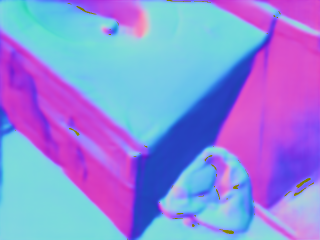}} &
{\includegraphics[width=0.18\linewidth]{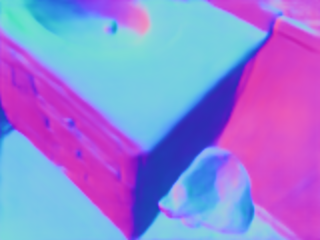}}
\\
{\includegraphics[width=0.18\linewidth]{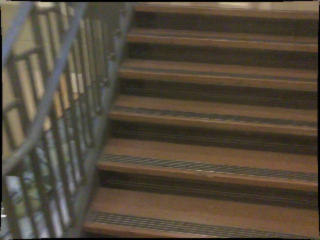}} & 
{\includegraphics[width=0.18\linewidth]{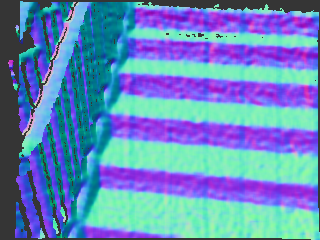}} &
{\includegraphics[width=0.18\linewidth]{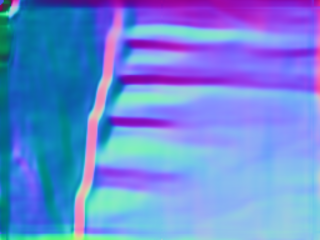}} &
{\includegraphics[width=0.18\linewidth]{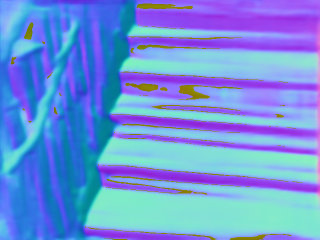}} &
{\includegraphics[width=0.18\linewidth]{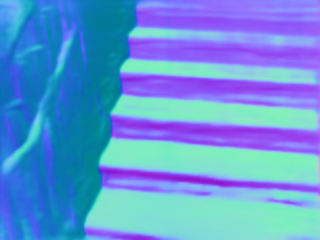}}
\\
{\includegraphics[width=0.18\linewidth]{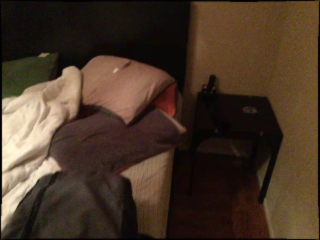}} & 
{\includegraphics[width=0.18\linewidth]{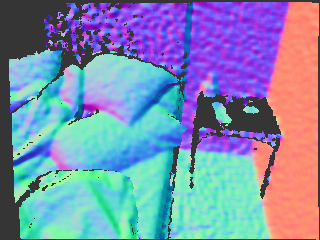}} &
{\includegraphics[width=0.18\linewidth]{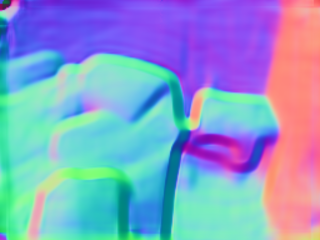}} &
{\includegraphics[width=0.18\linewidth]{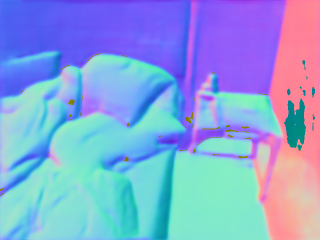}} &
{\includegraphics[width=0.18\linewidth]{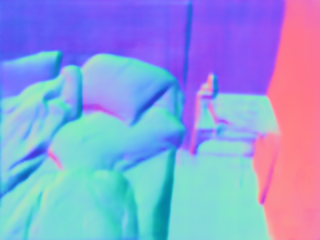}}
\\
Image & GT & CNM \cite{long2020occlusion} & NAS \cite{kusupati2020normal} & Ours \\
\end{tabular}
\centering
\vspace{0.05cm}
\caption{More qualitative results of surface normal estimation on ScanNet \cite{dai2017scannet}. \textbf{Better viewed when zoomed in.}}
\label{fig::supp_normal}
\vspace{-20pt}
\end{figure*}

We provide additional visualizations for both depth and surface normal estimation in Figure \ref{fig::supp_depth} and \ref{fig::supp_normal}. All samples are from the official test split of ScanNet \cite{dai2017scannet}.


\end{document}